\RequirePackage{fix-cm}
\documentclass[twocolumn]{svjour3}          
\smartqed  
\usepackage{bm}
\usepackage{dsfont}
\usepackage{times}
\usepackage{epsfig}
\usepackage{epstopdf}
\usepackage{amsmath}
\usepackage{amssymb}
\usepackage{subfigure}
\usepackage{caption}
\usepackage{tabularx}
\usepackage{graphicx}
\usepackage{wrapfig}
\usepackage{gensymb}
\usepackage[table]{xcolor}
\usepackage{cite} 
\usepackage{siunitx} 
\usepackage{textcomp}
 \usepackage[normalem]{ulem}
\usepackage[ruled,vlined,linesnumbered]{algorithm2e}
\usepackage[noend]{algpseudocode}
\usepackage[colorlinks,
            linkcolor=blue,
            anchorcolor=blue,
            citecolor=blue,
            bookmarks=false
            ]{hyperref}
\usepackage{booktabs}
\usepackage{multirow}
\usepackage[section]{placeins}
\makeatletter
\def\BState{\State\hskip-\ALG@thistlm}
\makeatother
\definecolor{myGreen}{HTML}{33FF00}
\definecolor{myRed}{HTML}{FF3030}
\definecolor{myGrey}{HTML}{AA5555}
\definecolor{myWhite}{HTML}{FFFFFF}
\definecolor{maroon}{cmyk}{0,0.87,0.68,0.32}
\definecolor{petr}{HTML}{5555FF}
\definecolor{josef}{HTML}{FF3030}

\journalname{International Journal of Computer Vision (IJCV)}
\begin{document}\sloppy

\title{Generalized 3D Self-supervised Learning Framework via Prompted Foreground-Aware Feature Contrast}





\author{Kangcheng Liu, Xinhu Zheng, Chaoqun Wang, Kai Tang, Ming Liu, and Baoquan Chen}

\maketitle

\begin{abstract}

Contrastive learning has recently demonstrated great potential for unsupervised pre-training in 3D scene understanding tasks. However, most existing work randomly selects point features as anchors while building contrast, leading to a clear bias toward background points that often dominate in 3D scenes. Also, object awareness and foreground-to-background discrimination are neglected, making contrastive learning less effective. To tackle these issues, we propose a general foreground-aware feature contrast FAC++ framework to learn more effective point cloud representations in pre-training. FAC++ consists of two novel contrast designs to construct more effective and informative contrast pairs. The first is building positive pairs within the same foreground segment where points tend to have the same semantics. The second is that we prevent over-discrimination between 3D segments/objects and encourage grouped foreground-to-background distinctions at the segment level with adaptive feature learning in a Siamese correspondence network, which adaptively learns feature correlations within and across point cloud views effectively. Moreover, we have designed the foreground-prompted regional sampling to enhance more balanced foreground-aware learning, which is termed FAC++. Visualization with point activation maps shows that our contrast pairs capture clear correspondences among foreground regions during pre-training. Quantitative experiments also show that FAC++ achieves superior knowledge transfer and data efficiency in various downstream 3D semantic segmentation, instance segmentation as well as object detection tasks. All codes, data, and models are available at:  \url{https://github.com/KangchengLiu/FAC_Foreground_Aware_Contrast}.
\vspace{-0.1001mm}


\vspace{-1.1001mm}

\end{abstract}
\vspace{-0.001mm}

\vspace{39.001mm}

\section{Introduction}
\label{intro}
\vspace{-0.001mm}
\vspace{-0.001mm}

    \begin{figure}[t]
        \centering
        \includegraphics[scale=0.46193]{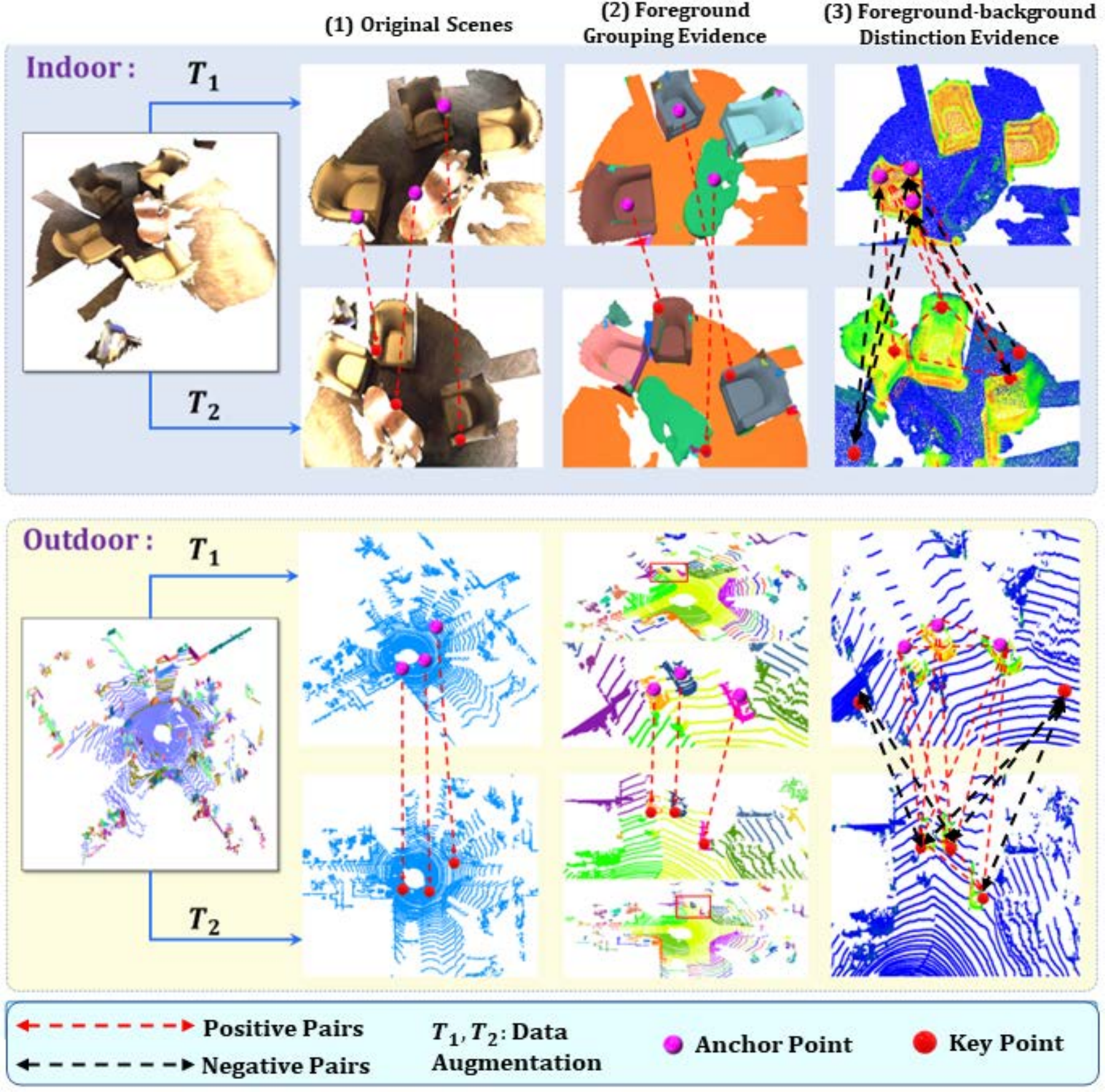}
        \caption{Construct informative contrast pairs matters in contrastive learning: Conventional contrast requires strict point-level correspondence. The proposed method FAC/FAC++ takes both the foreground grouping and the foreground-background distinction guidance into account, thus forming better contrast pairs to learn more informative and discriminative 3D feature representations.
        }     
        \label{fig_motivation}
    \end{figure}

3D scene understanding is crucial to many tasks such as robot grasping, smart manufacturing, virtual reality/augmented reality, and autonomous navigation~\cite{huang2018holistic, gal2021mrgan, liu2022less}. However, most existing work is fully supervised, which relies heavily on large-scale annotated 3D data that is often very laborious to collect. Self-supervised learning (SSL), which allows learning rich and meaningful representations from large-scale unannotated data, has recently demonstrated great potential to mitigate the annotation constraint~\cite{bar2022detreg, chen2020simple}. It learns with auxiliary supervision signals derived from unannotated data, which are usually much easier to collect. In particular, contrastive learning as one prevalent SSL approach has achieved great success in various visual downstream 2D recognition tasks~\cite{chen2021exploring, zbontar2021barlow, he2020momentum}.


Contrastive learning has also been explored for point cloud representation learning in various downstream tasks such as semantic segmentation~\cite{huang2021spatio, xie2020pointcontrast, chen2021shape, han2019multi}, instance segmentation~\cite{hou2021exploring, hou2021pri3d}, and object detection~\cite{liang2021exploring, yin2022proposalcontrast}. However, many successful 2D contrastive learning methods~\cite{chen2021exploring, zadaianchuk2022unsupervised, gokul2022refine} do not work well for 3D point clouds, largely because point clouds often capture wide-view scenes which consist of complex points of many irregularly distributed foreground objects as well as a large number of background points. Several studies attempt to design specific contrast to cater to the geometry and distribution of point clouds. For example, ~\cite{huang2021spatio} employs max-pooled features of two augmented scenes to form the contrast, but they tend to over-emphasize holistic information and overlook informative features about foreground objects. ~\cite{xie2020pointcontrast, hou2021exploring, liang2021exploring} directly use registered point/voxel features as positive pairs and treat all non-registered as negative pairs, causing many false contrast pairs in semantics.


We propose exploiting scene foreground evidence and foreground-background distinction to construct more \textit{foreground grouping aware} and \textit{foreground-background distinction aware} contrast for learning discriminative 3D representations. For \textit{foreground grouping aware} contrast, we first obtain regional correspondences with over-segmentation~\cite{papon2013voxel} and then build positive pairs with points of the same region across views, leading to semantic coherent representations. In addition, we design a sampling strategy to sample more foreground point features while building contrast, because the background point features are often less-informative and have repetitive or homogeneous patterns. For \textit{foreground-background distinction aware} contrast, we first enhance foreground-background point feature distinction and then design a Siamese correspondence network that selects correlated features by adaptively learning affinities among feature pairs within and across views in both foreground and background to avoid over-discrimination between parts/objects.  Visualizations show that, in a complementary manner, foreground-enhanced contrast effectively guides the learning toward foreground regions while foreground-background contrast enhances distinctions among foreground and background features. The two designs collaborates to learn more informative and discriminative representation as illustrated in Fig. \ref{fig_motivation}.

\vspace{-0.000001mm}


This work is a significant extension of the preliminary version of the published conference work~\cite{liu2022weaklyeccv}, where basic ideas of forming contrast between grouped foreground and background are introduced to tackle the final data-efficient 3D scene understanding during the fine-tuning stage to enhance the foreground grouping and foreground-background distinction. In summary, we extensively enriched previous works in the following aspects:

\vspace{-0.00001mm}


\textit{First,} we propose to enhance the foreground-background discrimination directly with explicit foreground points queries of FAC/FAC++ from the 3D vision-language models, which achieved more generalized pre-training for representation learning, and enhance the final data-efficient learning as well as the open-vocabulary recognition performance. 

\vspace{-0.000001mm}


\textit{Second,} we added the experiments in instance segmentation, where we have achieved superior performance for that the superior foreground-aware instance discrimination in feature representation learning is successfully achieved. 


\textit{Third,} we added comprehensive experimental about the open-world recognition, which demonstrate apart from state-of-the-art performance in data-efficient learning, our proposed approach is also compatible with the current state-of-the-art 3D open vocabulary recognition approaches, and also has superior effectiveness in terms of recognizing novel categories and novel semantic classes.


 \begin{figure*}[t]
        \centering  \includegraphics[scale=0.15593]{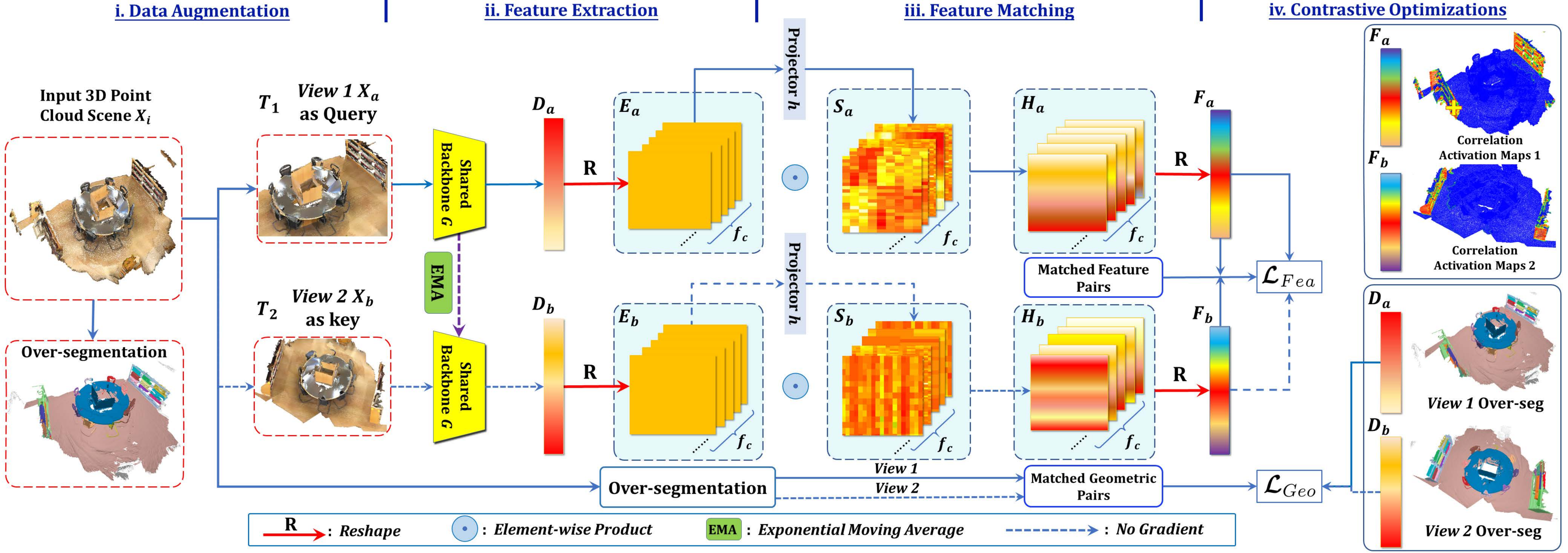}
        \caption{\textbf{The framework of our proposed FAC/FAC++.} FAC samples median-sized regions as foreground regions, while FAC++ leverages foreground prompts to enhance the foreground-aware feature representations. They both take two augmented 3D point cloud views as input which first extracts the backbone features $\textit{\textbf{D}}_a$ and $\textit{\textbf{D}}_b$ for foreground aware contrast with $\bm{\mathcal{L}}_{Geo}$. The backbone features are then reshaped to regularized representation $\textit{\textbf{E}}_a$ and $\textit{\textbf{E}}_b$ to find correspondences across two views for feature matching. Specifically, we adopt the projector $\textit{\textbf{h}}$ to transfer $\textit{\textbf{E}}_\textit{a}$ and $\textit{\textbf{E}}_\textit{b}$ to feature maps $\textit{\textbf{S}}_\textit{a}$ and $\textit{\textbf{S}}_\textit{b}$ to adaptively learn their correlations and produce enhanced representations $\textit{\textbf{H}}_a$ and $\textit{\textbf{H}}_b$. Finally, $\textit{\textbf{H}}_a$ and $\textit{\textbf{H}}_b$ are reshaped back to $\textit{\textbf{F}}_\textit{a}$ and $\textit{\textbf{F}}_\textit{b}$ where matched feature pairs are enhanced with feature contrast loss $\bm{\mathcal{L}}_{Fea}$. Hence, both our proposed FAC and FAC++ exploits complementary foreground awareness and foreground-background distinction within and across views for more informative and meaningful representation learning.
        }
        \label{fig_frame}
\end{figure*}



The contributions of this work can be summarized in the following aspects. \textit{First,} we propose FAC/FAC++, a foreground-aware feature contrast framework for large-scale 3D pre-training. FAC samples median-sized regions as foreground regions, while FAC++ leverages foreground prompts to enhance the foreground-aware feature representations. \textit{Second,} we construct region-level contrast to enhance the local coherence and better foreground awareness in the learned representations. \textit{Third,} on top of that, we design a Siamese correspondence framework that can locate well-matched keys to adaptively enhance the intra- and inter-view feature correlations, as well as enhance the foreground-background distinction. \textit{Fourthly,} we propose leveraging current prevailing vision-language models to extend the model's generalization capacity while encountered with novel categories, and demonstrate the open-world recognition capacity of the model by extensive experiments. \textit{Lastly,} extensive experiments over multiple public benchmarks show that FAC++ achieves superior self-supervised learning when compared with the state-of-the-art. FAC++ is compatible with the prevalent 3D segmentation backbone network SparseConv~\cite{graham20183d} and 3D detection backbone networks including PV-RCNN~\cite{shi2020pv}, PointPillars~\cite{lang2019pointpillars}, and Point-RCNN~\cite{shi2019pointrcnn}. It is also applicable to both indoor dense RGB-D and outdoor sparse LiDAR point clouds. 

\vspace{-0.21001mm}
\section{Related Work}
\vspace{-0.001mm}
\vspace{-0.001mm}
\vspace{-2.001mm}

\subsection{3D Scene Understanding}
\vspace{-2.001mm}

3D scene understanding aims for understanding of 3D depth or point-cloud data and it involves several downstream tasks such as 3D semantic segmentation~\cite{chibane2022box2mask, liu2021contrastive}, 3D object detection~\cite{yin2021center, erccelik20223d}, etc. It has recently achieved very impressive progress as driven by the advance in 3D deep learning strategy and the increasing large-scale 3D benchmark datasets. Different approaches have been proposed to address various challenges in 3D scene understanding. For example, the point-based approach~\cite{liu2022weaklyeccv, hu2020randla, zhao2021point, park2022fast, xu2021paconv} can learn point features well but is often stuck by high computational costs while facing large-scale point-cloud input stream. Voxel-based approach~\cite{choy20194d, mao2021voxel, zhou2018voxelnet, liu2022integratedarxiv, jiang2021guided, liu2022weaklabel3d, liu2021one} is computation and memory efficient but often suffers from information loss from the voxel quantification. In addition, voxel-based SparseConv network~\cite{vu2022softgroup, graham20183d, choy20194d} has shown very promising performance in indoor 3D scene parsing, while combining point and voxel often has a clear advantage in outdoor LiDAR-based detection~\cite{lang2019pointpillars, zhou2018voxelnet, shi2020points, shi2020pv}. Our proposed SSL framework shows consistent superiority in indoor/outdoor 3D perception tasks, and it is also backbone-agnostic.
\vspace{-0.291mm}

\subsection{Self-supervised Pre-training on Point Clouds}
\vspace{-0.2mm}

\noindent
\textbf{Contrastive Pre-training.} Recent years have witnessed notable success in contrastive learning for learning unsupervised representations~\cite{li2022closer, wang2021unsupervised, zhang2022self, sanghi2020info3d}. For example, contrast scene context (CSC)~\cite{hou2021exploring, xie2020pointcontrast} explores contrastive pre-training with scene context descriptors. However, it focuses too much on optimizing low-level registered point features but neglects the regional homogeneous semantic patterns and high-level feature correlations. Some work employs max-pooled scene-level information for contrast~\cite{huang2021spatio, liang2021exploring,zhang2021self}, but it tends to sacrifice local geometry details and object-level semantic correlations, leading to sub-optimal representations for dense prediction tasks such as semantic segmentation. Differently, we explicitly consider regional foreground awareness as well as feature correlation and distinction among foreground and background regions which lead to more informative and discriminative representations in 3D downstream tasks. 

Further, many approaches incorporate auxiliary temporal or spatial 3D information for self-supervised contrast with augmented unlabeled datasets~\cite{huang2021spatio} and synthetic CAD models~\cite{wu20153d, yamada2022point,rao2021randomrooms,chen20224dcontrast}.  STRL~\cite{huang2021spatio} introduces a mechanism of learning from dynamic 3D scenes synthetic 3D by regarding 3D scenes are RGB-D video sequences. 
Randomrooms~\cite{rao2021randomrooms} synthesizes man-made 3D scenes by randomly putting synthetic CAD models into regular synthetic 3D scenes. 4DContrast~\cite{chen20224dcontrast} leverages spatio-temporal motion priors of synthetic 3D shapes to learn a better 3D representation. However, most of these prior studies rely on extra supervision from auxiliary spatio-temporal information. Differently, we perform self-supervised learning over original 3D scans without additional synthetic 3D models.

\noindent    
\textbf{Masked Generation-based Pre-training.} Masked image modeling has demonstrated its effectiveness in various image understanding tasks~\cite{xie2022simmim, he2022masked} with the success of vision transformers~\cite{liu2021swin, carion2020end}. Recently, mask-based pre-training~\cite{liang2022meshmae, pang2022masked, rao2020global, han2019view, wang2021unsupervised} has also been explored for the understanding of small-scale 3D shapes~\cite{wu20153d, uy2019revisiting, chang2015shapenet}. However, mask-based designs usually involve a transformer backbone~\cite{liang2022meshmae, pang2022masked} that has a high demand for both computation and memory while handling large-scale 3D scenes. We focus on pre-training with contrastive learning, which is compatible with both point-based and voxel-based backbones.  

\vspace{-0.02mm}

\section{Method}
As illustrated in Fig.~\ref{fig_frame}, our proposed FAC/FAC++ frameworks are composed of four components: data augmentation, backbone network feature extraction, feature matching, and foreground-background aware feature contrastive optimizations with matched contrast pairs. The differences of them merely lie in that FAC samples median-sized regions as foreground regions, while FAC++ leverages foreground prompts to enhance the foreground-aware feature representations. In the following, we first revisit typical contrastive learning approaches for 3D point clouds and discuss their limitations that could lead to less informative representations. We then elaborate our proposed FAC from three major aspects: 1) Regional grouping contrast that exploits local geometry homogeneity from over-segmentation to encourage semantic coherence of local regions; 2) A correspondence framework that consists of a Siamese network and a feature contrast loss for capturing the correlations among the learned feature representations; 3) Optimization losses that take advantage of the better contrast pairs for more discriminative self-supervised learning.

\subsection{Point- and Scene-Level Contrast Revisited}


The key in contrastive learning-based 3D SSL is to construct meaningful contrast pairs between the two augmented views. Positive pairs have been constructed at either point level as in PointContrast (PCon)~\cite{xie2020pointcontrast} or scene level as in DepthContrast (DCon)~\cite{zhang2021self}. Concretely, given the augmented views of 3D partial point/depth scans, the contrastive loss is applied to maximize the similarity of the positive pairs and the distinction between negative pairs. In most cases, InfoNCE~\cite{oord2018representation} loss can be applied for contrast:  
\vspace{-0.211mm}

    \begin{scriptsize}
    \begin{equation}
    \mathcal{L}_{ctra}=-\frac{1}{\|\textbf{B}_p\|}\sum_{(a, b) \in \textbf{B}_p}\log\frac{\exp(\textit{\textbf{f}}_{g1}^{a} \cdot \textit{\textbf{f}}_{g2}^{b}/\tau)}{\sum_{(\cdot, c) \in \textbf{B}_p} \exp(\textit{\textbf{f}}_{g1}^{a} \cdot \textit{\textbf{f}}_{g2}^{c}/\tau))}.
    \end{equation}
    \end{scriptsize}
    
\noindent

Here $\textit{\textbf{f}}_{g1}$ and $\textit{\textbf{f}}_{g2}$ are the feature vectors of two augmented views for contrast. $\textbf{B}_p$ is the index set of matched positive pairs. $(a, b) \in \textbf{B}_p$ is a positive pair whose feature embeddings are forced to be similar, while $\{(a, c)|(\cdot, c)\in \textbf{B}_p, c \neq b \}$ are negative pairs whose feature embeddings are encouraged to be different. PCon~\cite{xie2020pointcontrast} directly adopts registered point-level pairs while DCon~\cite{zhang2021self} uses the max-pooled scene-level feature pairs for conducting contrast.

\vspace{-0.001 mm}

Despite their decent performance in 3D downstream tasks, the constructed contrast pairs in prior studies tend to be sub-optimal. As illustrated in Fig.~\ref{fig_motivation}, point-level contrast tends to overemphasize the fine-grained low-level details and overlook the region-level geometric coherence which often provides object-level information. Scene-level contrast aggregates the feature of the whole scene for contrast, which can lose the object-level spatial contexts and distinctive features, leading to less informative representations for downstream tasks. We thus conjecture that region-level correspondences are more suitable to form the contrast, and this has been experimentally verified as illustrated in Fig.~\ref{fig_motivation}, more details to be elaborated in ensuing Subsections. 
\vspace{-0.001mm}

\vspace{-0.001mm}



\subsection{Foreground-Aware Contrast}
\label{Subsect:Geo}
   Region-wise feature representations have been shown to be very useful in considering contexts for downstream tasks such as semantic segmentation and detection~\cite{he2017mask, zhang2020h3dnet, bai2022point}. In our proposed geometric region-level foreground-aware contrast, we obtain regions by leveraging the off-the-shelf point cloud over-segmentation techniques~\cite{papon2013voxel, guo20143d}. The adoption of over-segmentation is motivated by its merits in three major aspects.  First, it can work in a completely unsupervised manner without requiring any annotated data. Second, our proposed regional sampling (to be described later) allows us to filter out background regions such as ceilings, walls, and ground in an unsupervised manner, where the background regions are often represented by geometrically homogeneous patterns with a large number of points. Regions with a very limited number of points can also be filtered out, which are noisy in both geometry and semantics. Third, over-segmentation provides geometrically coherent regions with high semantic similarity, while diverse distant regions tend to be semantically distinct after sampling, which effectively facilitates discriminative feature learning. Specifically, over-segmentation divides the original point clouds scene into $I$ class-agnostic regions $S=\{s_1, s_2,..., s_i\}$, and ${s_{i}} \cap {s_{k}}=\emptyset$ for any ${s_{i}} \neq {s_{k}}$.  Our empirical experiments show that our proposed framework works effectively with mainstream over-segmentation approaches without fine-tuning. 


\noindent
\textbf{Foreground Prompted Regional Sampling for Balanced Learning.}
    In our preliminary conference version, we designed a simple but effective region sampling technique to obtain meaningful foreground from the geometrically homogeneous regions as derived via over-segmentation as introduced above. Specifically, we first count the number of points in each region and rank regions according to the number of points they contain. We then identify the region having the median number of points as $s_{med}$. Next, we select $H$ regions having the closest number of points with $s_{med}$ to form contrast pairs. Extensive experiments show that this sampling strategy is effective in the downstream task. We conjecture that the massive points in background regions encourage biased learning towards repetitive and redundant information, while regions with very limited points are noisy in both geometry and semantics. Our sampling strategy can encourage balanced learning towards foreground regions which leads to more informative and discriminative representations.


\noindent
   \textbf{Foreground Prompts.} Note that for FAC, merely selecting the median-sized regions can not always guarantee that all points lie within this region are foreground points. Therefore, we designed an approach to better guide the contrastive optimizations to more effectively enhance the foreground-background distinctive representations. Specifically, we designed an effective approach for the language-queried foreground sampling based on the current prevailing vision-language models~\cite{bai2023qwen}. We directly utilize the model from the OpenMask3D~\cite{takmaz2023openmask3d} to obtain the aligned 3D and language co-embeddings. Then, we utilize the pre-trained models with the corresponding language textual query termed "foreground" to obtain the final filtered regions with foreground points. It tuns out that this simple but effective design filters out many non-foreground points and provide more clearly separate foreground regions, which will be of significance to the final downstream 3D scene understanding tasks.


\noindent
\textbf{Contrast with Local Regional Consistency.}
Different from the above-mentioned PCon \cite{xie2020pointcontrast} and DCon \cite{zhang2021self}, we directly exploit region homogeneity to obtain contrast pairs. Specifically, taking the average point feature within a region as the anchor, we regard selected features within the same region as positive keys and in different regions as negative keys. Benefiting from the region sampling strategy, we can focus on the foreground for better representation learning. Denote the number of points within a region as $\mathcal{N}(s_i)$ and the backbone feature as $\textit{\textbf{D}}$, we aggregated their point feature $\textit{\textbf{d}}_j \in \textit{\textbf{D}}$ to produce an average regional feature $\bm{\mathcal{D}}_m$ within a region as the anchor in contrast to enhance the robustness:

\vspace{-5mm}
     \begin{center} 
    \begin{scriptsize}
    \centering
        \begin{equation}
         \bm{\mathcal{D}}_m=\frac{1}{\|\mathcal{N}(s_i)\|}\sum_{{j\in 
         \mathcal{N}(s_i)}} \textit{\textbf{d}}_j.
        \end{equation}
    \end{scriptsize}
    \end{center}

 \noindent Regarding $\bm{\mathcal{D}}_m$ as the anchor, we propose a foreground aware geometry contrast loss $\bm{\mathcal{L}}_{Geo}$ pulling the point feature to its corresponding positive features in the local geometric region, and pushing it apart from negative point features of different separated regions: 
 
    
    \begin{scriptsize}
        \begin{equation}
        \bm{\mathcal{L}}_{Geo}=-\frac{1}{\|\textbf{B}_p\|}\sum_{(a, b) \in \textbf{B}_p}\log\frac{\exp(\bm{\mathcal{D}}^a_m \cdot \textit{\textbf{d}}_{j}^{b, +} /\tau)}{\sum_{(\cdot, c) \in \textbf{B}_p} \exp(\bm{\mathcal{D}}^a_m \cdot \textit{\textbf{d}}_{j}^{b, -} /\tau))}.
        \end{equation}
    \end{scriptsize}
      
 \noindent Here, $\textit{\textbf{d}}_{j}^{+}$ and $\textit{\textbf{d}}_{j}^{-}$ denote the positive and negative samples, respectively with $\bm{\mathcal{D}}_m$. We set the number of positive and negative point feature pairs for each regional anchor as $k$ equally.  
Note our proposed foreground contrast is a generalized version of PCon \cite{xie2020pointcontrast} with foreground enhanced and it returns to PCon if all regions shrink to a single point. Benefiting from the regional geometric consistency and balanced foreground sampling, the foreground aware contrast alone outperforms the state-of-the-art CSC \cite{hou2021exploring} in data efficiency in our empirical experimental results.

\noindent
\subsection{Foreground-Background Distinction Aware Contrast}

As illustrated in Fig.~\ref{fig_frame}, we propose a Siamese correspondence network (SCN) to explicitly identify feature correspondences within and across views and introduce a feature contrast loss to adaptively enhance their correlations. The SCN is merely used during the pre-training stage for improving the representation quality. After pre-training, only the backbone network is fine-tuned for downstream tasks. \\
\noindent
\textbf{Siamese Correspondence Network for Adaptive Correlation Mining.} 
Given the input 3D scene $\textit{\textbf{X}}_i$ with $N$ points, our proposed FAC/FAC++ framework first transforms it to two augmented views $\textit{\textbf{X}}_a$ and $\textit{\textbf{X}}_b \in \mathbb{R}^{N \times {f_{in}}}$, and obtain backbone feature $\textit{\textbf{D}}_a$ and $\textit{\textbf{D}}_b \in \mathbb{R}^{N \times {f_{c}}}$ by feeding the two views into the backbone network $\mathcal{G}$ and its momentum update (via exponential moving average), respectively ($f_{c}$ is the number of feature channels).  For fair comparisons, we adopt the same augmentation scheme with existing work~\cite{yin2022proposalcontrast, hou2021exploring}. In addition, We reshape the backbone point-level features to feature maps $\textit{\textbf{E}}_a$ and $\textit{\textbf{E}}_b \in \mathbb{R}^{m \times \frac{N}{m} \times {f_{c}}}$ to obtain regularized point cloud representations and reduce computational costs. We then apply the projector $\textit{\textbf{h}}$ to  $\textit{\textbf{E}}_a $ and $\textit{\textbf{E}}_b $ respectively to obtain feature maps $\textit{\textbf{S}}_\textit{a}$ and $\textit{\textbf{S}}_\textit{b} \in \mathbb{R}^{\textit{m} \times \frac{N}{\textit{m}} \times {f_{c}}}$ of the same dimension as $\textit{\textbf{E}}_a$ and $\textit{\textbf{E}}_b$. We adopt two simple point-MLPs with a ReLU layer in between to form the projector $\textit{\textbf{h}}$. The feature maps $\textit{\textbf{S}}_\textit{a}$ and $\textit{\textbf{S}}_\textit{b}$ work as learnable scores which adaptively enhance the significant and correlated features within and across two views. Finally, we conduct element-wise product between $\textit{\textbf{E}}$ and $\textit{\textbf{S}}$ to obtain the enhanced feature $\textit{\textbf{H}}_\textit{a}$ and $\textit{\textbf{H}}_\textit{b} \in \mathbb{R}^{m \times \frac{N}{m} \times {f_{c}}}$ and further transform them back to point-wise features $\textit{\textbf{F}}_a$ and $\textit{\textbf{F}}_b \in \mathbb{R}^{N \times {f_{c}}}$ for correspondence mining. The global feature-level discriminative representation learning is enhanced by the proposed SCN, enabling subsequent contrast with the matched feature.


\noindent
\textbf{Contrast with the Matched Feature and Foreground-Background Distinction.} 
With the obtained sampled foreground-background pairs labeled as negative, we conduct feature matching to select the most correlated positive contrastive pairs. As illustrated in Fig.~\ref{fig_frame}, we evaluate the similarity between $\textit{\textbf{F}}_a$ and $\textit{\textbf{F}}_b$ and select the most correlated pairs for contrast. The regional anchors are selected in the same manner as in Subsection~\ref{Subsect:Geo}. Concretely, we first introduce an average feature $\bm{\mathcal{F}}_m^a$ for point feature within a region as the anchor when forming contrast, given as $\bm{\mathcal{F}}_m^a=\frac{1}{\|\mathcal{N}(s_i)\|}\sum_{{j\in 
\mathcal{N}(s_i)}} \textbf{\textit{f}}_j^a$, based on the observation that points in the same local region tends to have the same semantic. For \textit{j-th} point-level feature $\textit{\textbf{f}}^{\;b}_j \in \mathbb{R}^{c}$ in $\textit{\textbf{F}}_b$, we calculate  its similarity $\textit{S}_{p,j}$ with regional feature $\bm{\mathcal{F}}_m^a \in \mathbb{R}^{c}$:

    \begin{scriptsize}
\begin{equation}
     \textit{S}_{p, j}= \mathcal{F_S}(\bm{\mathcal{F}}_m^a, \textit{\textbf{f}}^{\;b}_j).
\end{equation}
    \end{scriptsize}


\begin{figure}[t]
    \centering
    \includegraphics[scale=0.241]{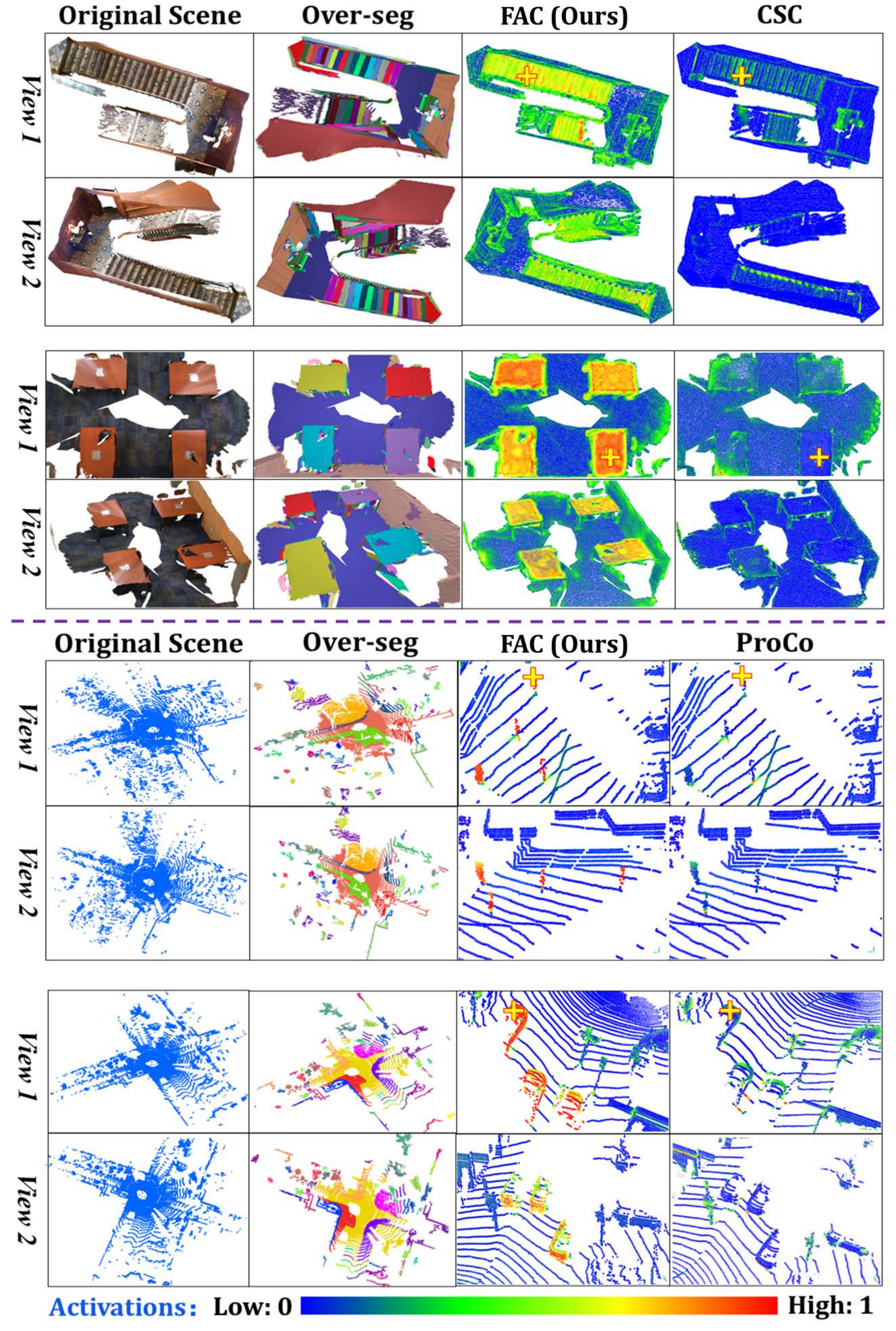}
    \caption{ Visualizations of projected point correlation maps over the indoor ScanNet (1st-4th rows) and the outdoor KITTI \cite{geiger2013vision} (5th-8th rows) with respect to the query points highlighted by yellow crosses. The \textit{View 1} and \textit{View 2} in each sample show the intra-view and cross-view correlations, respectively. We compare FAC with the state-of-the-art CSC \cite{hou2021exploring} on segmentation (rows 1-4) and ProCo \cite{yin2022proposalcontrast} on detection (rows 5-8). FAC clearly captures better feature correlations within and across views (columns 3-4).
    }
    \label{fig_activation}
    \vspace{-0.0001mm}
\end{figure}

Here $\mathcal{F_S}(\textit{\textbf{x}}, \textit{\textbf{y}})$ denotes the cosine similarity between vectors $\textit{\textbf{x}}$ and $\textit{\textbf{y}}$. We sample the top-$k$ elements from $\textit{S}_{p, j}$ as positive keys with the regional feature $\bm{\mathcal{F}}_m^a$ from both foreground and background point features. The top-$k$ operation is easily made differentiable by reformulating it as an optimal transport problem. Besides, we equally select other $k$  foreground-background pairs as negative.

\begin{scriptsize}
\begin{equation}
\bm{\mathcal{L}}_{Fea}=-\frac{1}{\|\textbf{B}_p\|}\sum_{(a, b) \in \textbf{B}_p}\log\frac{\exp(\bm{\mathcal{F}}_m^a  \cdot \textit{\textbf{f}}^{\;b,+}_{j} /\tau)}{\sum_{(\cdot, c) \in \textbf{B}_p} \exp(\bm{\mathcal{F}}_m^a \cdot \textit{\textbf{f}}^{\;c,-}_{j}  /\tau))}.
\end{equation}
\end{scriptsize}



Here, $\textit{\textbf{f}}^{\;b,+}_{j}$ denotes the positive keys of the identified \textit{k} most similar elements with $\bm{\mathcal{F}}_m^a$ from $\textit{\textbf{F}}_b$ in another view. $\textit{\textbf{f}}^{\;b,-}_{j}$ denotes the sampled other $k$ negative point features in a batch, respectively. Therefore, the well-related cross-view point features can be adaptively enhanced with the learning of the point-level feature maps $\textit{\textbf{S}}_\textit{a}$ and $\textit{\textbf{S}}_\textit{b}$ of 3D scenes. Our feature contrast enhances the correlations at the feature level within and across views by explicitly finding the region-to-point most correlated keys for the foreground anchor as the query. With learned feature maps, the features of well-correlated foreground/background points are adaptively emphasized while foreground-background distinctive ones are suppressed. Our proposed framework is verified to be very effective qualitatively in point activation maps and quantitatively in downstream transfer learning and data efficiency.



\subsection{Joint Optimization of Our Framework}

Considering both local region-level foreground geometric correspondence and global foreground-background distinction within and across views, the overall objective function of FAC/FAC++ framework $\mathcal{L}_{Sum}$ is as follows:
\vspace{-2.01mm}


\begin{equation}
\mathcal{L}_{Sum}=\alpha \mathcal{L}_{Geo} + \beta \mathcal{L}_{Fea}.
\end{equation}

\begin{figure*}[t!]
    \centering
    \includegraphics[scale=0.139999]{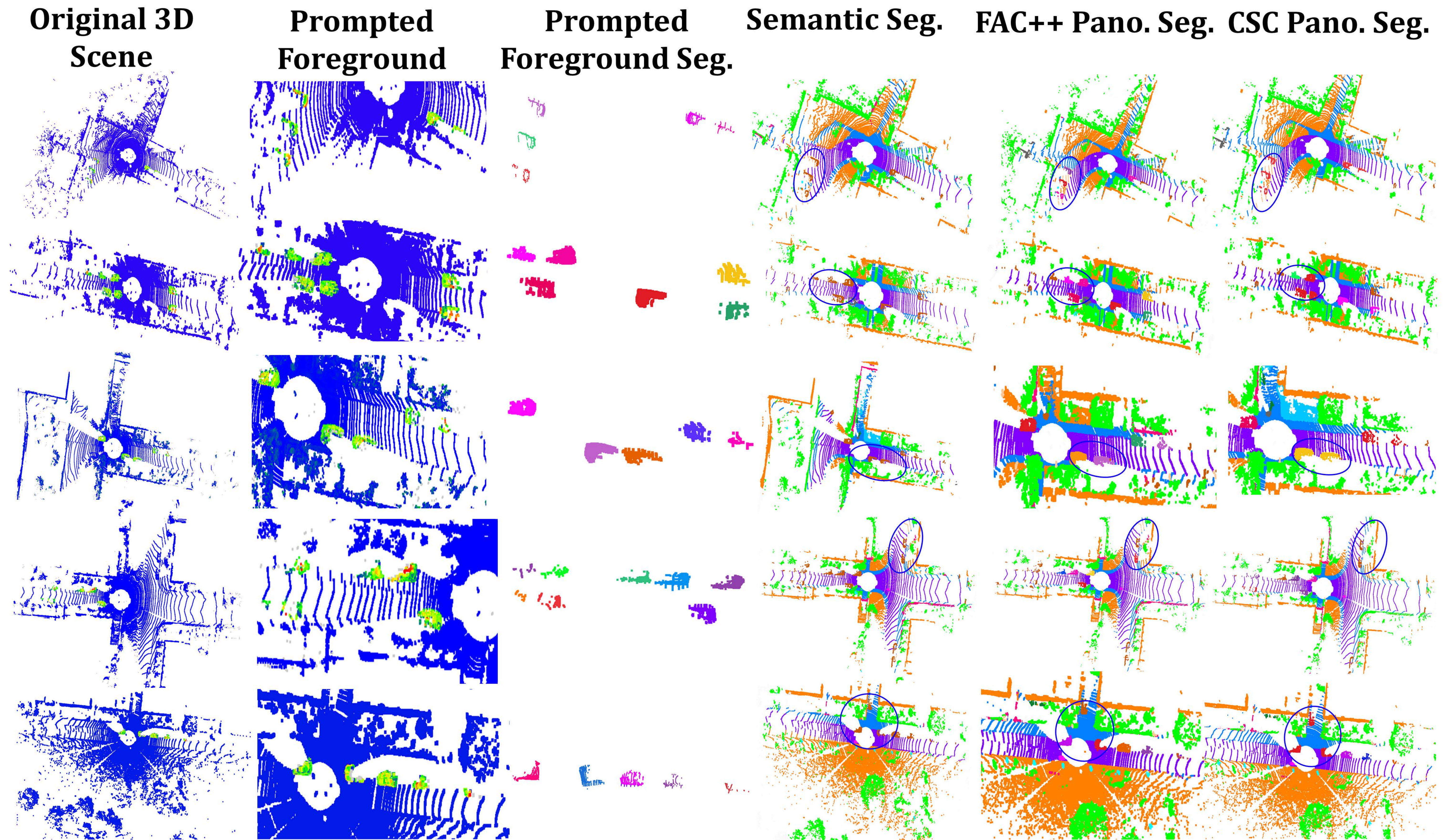}
    \caption{ Visualizations of the outdoor panoptic segmentation in SemanticKITTI. It is demonstrated that our proposed language queries can provide explicit foreground regional information, and the final panoptic segmentation performance is qualitatively superior, which successfully separates between diverse foreground objects very explicitly.
    }
    \label{fig_Vis_pano}
    \vspace{-1.16mm}
\end{figure*}


\vspace{-0.2mm}

Here $\alpha, \beta$ are the weights balancing two loss terms. We empirically set $\alpha=\beta=1$ without tuning.

\vspace{-0.18mm}
\vspace{-8.688888mm}


\section{Experiments}
\vspace{-1.08mm}

Data-efficient learning and knowledge transfer capacity have been widely adopted for evaluating self-supervised pre-training and the learned unsupervised representations \cite{hou2021exploring}. In the following experiments, we first pre-train models on large-scale unlabeled data and then fine-tune them with small amounts of labeled data of downstream tasks to test their data efficiency. We also transfer the pre-trained models to other datasets to evaluate their knowledge transfer capacity. The two aspects are evaluated over multiple downstream tasks including 3D semantic segmentation, instances segmentation, and object detection. Details of the involved datasets are provided in the Appendix.

\begin{figure}[t!]
    \centering
    \includegraphics[scale=0.244]{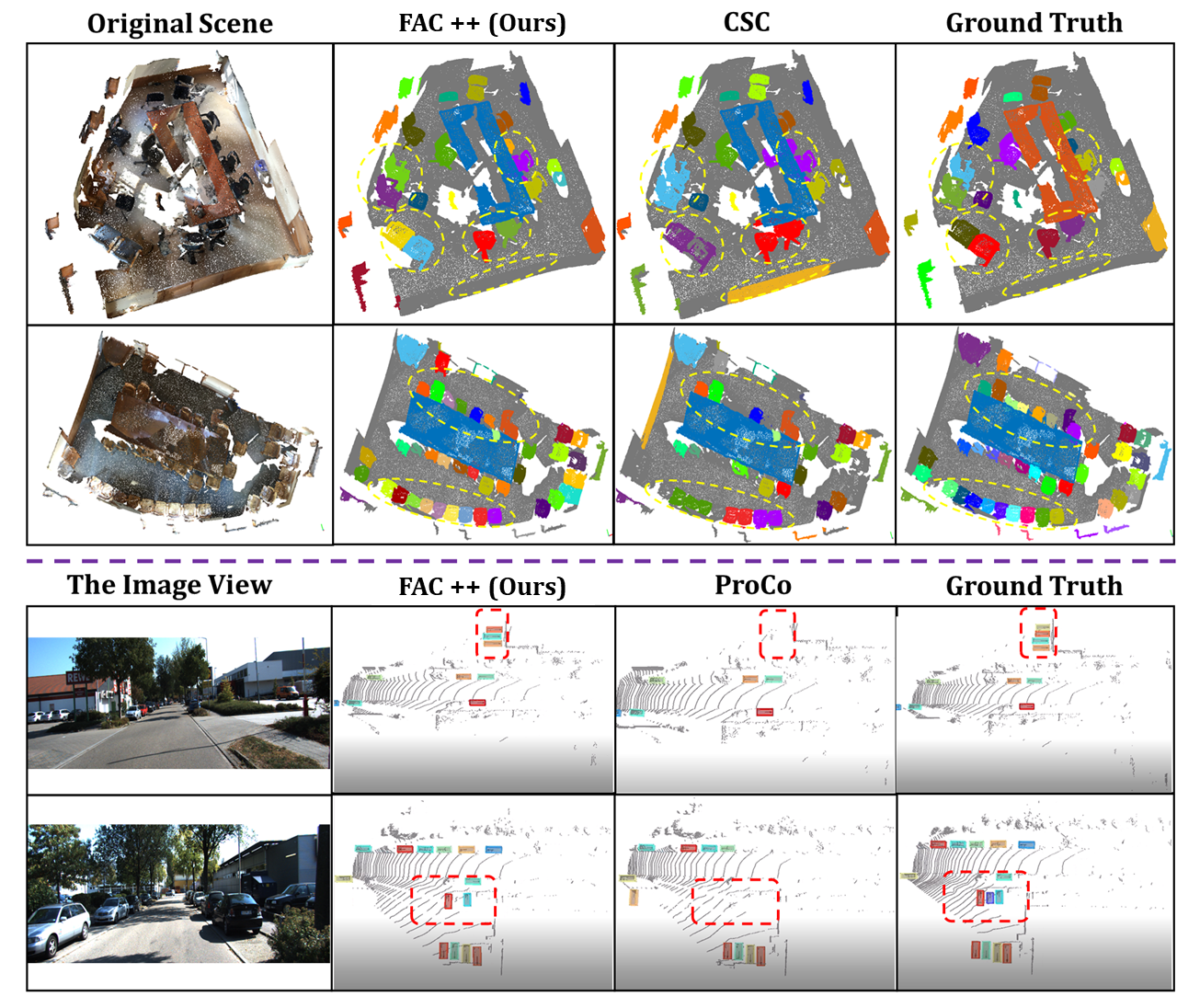}
    \caption{ Visualizations of indoor 3D segmentation over ScanNet compared with CSC \cite{hou2021exploring} as fine-tuned with 10\% labeled training data and outdoor object detection over KITTI \cite{geiger2013vision} with 20\% labeled training data compared with ProCo~\cite{yin2022proposalcontrast}. Different segmented instances and detected objects are highlighted in different colors. Differences in prediction are highlighted with the yellow ellipses as well as the red boxes.
    }
    \label{fig_Vis_3D}
    \vspace{-7.18mm}
\end{figure}

\vspace{-6.9mm}

\subsection{Experimental Settings}
\vspace{-1.9mm}

\noindent 
\textbf{3D Object Detection.} The object detection experiments involve two backbones including VoxelNet~\cite{zhou2018voxelnet} and PointPillars~\cite{lang2019pointpillars}.  Following ProCo~\cite{yin2022proposalcontrast}, we pre-train the model on Waymo and fine-tune it on KITTI~\cite{geiger2013vision} and Waymo.

\begin{figure*}[t!]
    \setlength{\abovecaptionskip}{-0.008cm}
    \setlength{\belowcaptionskip}{-0.008cm}
    \centering
\includegraphics[scale=0.1596888]{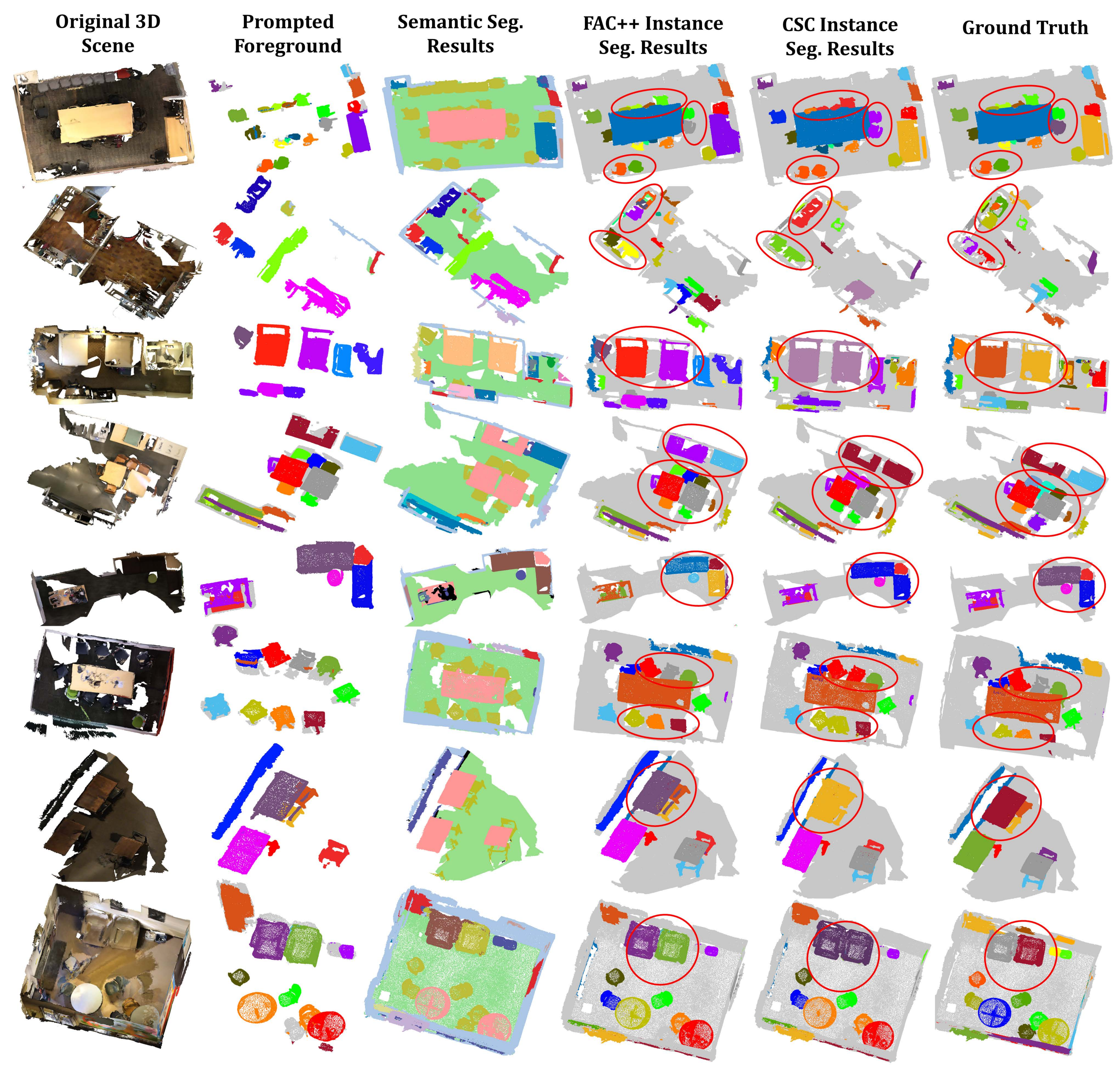}
    \caption{Visualizations of indoor 3D segmentation over ScanNet compared with CSC~\cite{hou2021exploring} as fine-tuned with 10\% labeled training data and outdoor object detection over KITTI~\cite{geiger2013vision} with 20\% labeled training data compared with ProCo~\cite{yin2022proposalcontrast}. It can also be demonstrated that our proposed languge query provides explicit as well as clearly separated foreground regional information. Different segmented instances and detected objects are highlighted with different colors. Differences in prediction are highlighted with yellow ellipses and red boxes.
    }
    \label{fig_Vis_3D_final}
    \vspace{-0.18mm}
\end{figure*}

\begin{table*}[htbp]
     \setlength{\abovecaptionskip}{-0.008cm}
\centering
\resizebox{\linewidth}{!}{
    \begin{tabular}{l||l||l||l||ll||ll||ll||ll}
     \toprule [0.5mm]
     \textbf{Fine-tuning with} & \multirow{2}{*}{\textbf{3D Detector}}& \multirow{2}{*}{\textbf{Approach}} & \multirow{1}{*}{\textbf{Pre-training}} & \multicolumn{2}{c||}{\textbf{Overall}} & \multicolumn{2}{c||}{\textbf{Vehicle}}  & \multicolumn{2}{c||}{\textbf{Pedestrian}} & \multicolumn{2}{c}{\textbf{Cyclist}}\\
     \textbf{Different Label Ratios} & & & \textbf{Schedule} & \textbf{AP\%} & \textbf{APH\%}&  \textbf{AP\%} & \textbf{APH\%} & \textbf{AP\%} & \textbf{APH\%} & \textbf{AP\%} & \textbf{APH\%}
     \\
     
     \midrule
     \multirow{8}{*}{1\% (around 0.8k frames)} & \multirow{4}{*}{PointPillars \cite{lang2019pointpillars}} & - & \textit{From Scratch} & 23.05 & 18.08 & 27.15 & 26.17 & 30.31 & 18.79 & 11.28 & 9.28 \\
     
     \multirow{3}{*}{} & & ProCo \cite{yin2022proposalcontrast}  & Pre-trained & 31.65 & 26.34 & 35.88 & 35.08 & 37.61 & 25.22 & 21.47 & 18.73 \\
     
     & & \textbf{FAC (Ours)} \cellcolor[RGB]{238,238,238} & Pre-trained \cellcolor[RGB]{238,238,238}
     & \textbf{33.57} 
     \cellcolor[RGB]{238,238,238} 
     & \textbf{28.13} 
     
     \cellcolor[RGB]{238,238,238} 
     & \textbf{37.92}
     \cellcolor[RGB]{238,238,238} 
     & \textbf{36.59} 
     \cellcolor[RGB]{238,238,238} 
     & \textbf{39.22} 
     
     \cellcolor[RGB]{238,238,238}
     & \textbf{26.78}
     
     \cellcolor[RGB]{238,238,238}  
     & \textbf{23.02} 
     
     
     \cellcolor[RGB]{238,238,238}
     & \textbf{20.22} 
     
     
     \cellcolor[RGB]{238,238,238}\\

      & & \textbf{FAC++ (Ours)} \cellcolor[RGB]{238,238,238} & Pre-trained \cellcolor[RGB]{238,238,238}
     & \textbf{34.58} 
     \cellcolor[RGB]{238,238,238} 
     & \textbf{29.22} 
     
     \cellcolor[RGB]{238,238,238} 
     & \textbf{38.86}
     \cellcolor[RGB]{238,238,238} 
     & \textbf{37.62} 
     \cellcolor[RGB]{238,238,238} 
     & \textbf{39.33} 
     
     \cellcolor[RGB]{238,238,238}
     & \textbf{26.89}
     
     \cellcolor[RGB]{238,238,238}  
     & \textbf{23.63} 
     
     
     \cellcolor[RGB]{238,238,238}
     & \textbf{21.36} 
     
     
     \cellcolor[RGB]{238,238,238}\\
     \cline{2-12}

      
     &\multirow{4}{*}{VoxelNet \cite{zhou2018voxelnet}} & - & \textit{From Scratch} & 20.88 & 17.83 & 21.95 & 21.45 & 27.98 & 20.52 & 12.70 & 11.53\\
     & & ProCo 
     \cite{yin2022proposalcontrast} & Pre-trained &38.36 & 34.78 & 37.60 & 36.91 & 39.74 & 31.70 &37.74 & 35.73\\
     & & \textbf{FAC (Ours)} \cellcolor[RGB]{238,238,238}& Pre-trained  \cellcolor[RGB]{238,238,238}
     & \textbf{40.15}


     \cellcolor[RGB]{238,238,238} 
     & \textbf{36.65} 
     

     \cellcolor[RGB]{238,238,238} & \textbf{39.57} 
     
     
     \cellcolor[RGB]{238,238,238} & \textbf{38.76} 
     
     
     \cellcolor[RGB]{238,238,238} & \textbf{41.59}  
     
     
     \cellcolor[RGB]{238,238,238} & \textbf{33.42} 
     
     
     \cellcolor[RGB]{238,238,238} & \textbf{39.43}
     
     
     \cellcolor[RGB]{238,238,238} & \textbf{37.39} 

\\

 & & \textbf{FAC++ (Ours)} \cellcolor[RGB]{238,238,238} & Pre-trained \cellcolor[RGB]{238,238,238}
     & \textbf{41.69} 
     \cellcolor[RGB]{238,238,238} 
     & \textbf{37.63} 
     
     \cellcolor[RGB]{238,238,238} 
     & \textbf{40.76}
     \cellcolor[RGB]{238,238,238} 
     & \textbf{39.79} 
     \cellcolor[RGB]{238,238,238} 
     & \textbf{42.53} 
     
     \cellcolor[RGB]{238,238,238}
     & \textbf{34.89}
     
     \cellcolor[RGB]{238,238,238}  
     & \textbf{40.93} 
     
     
     \cellcolor[RGB]{238,238,238}
     & \textbf{41.36} 
     
     
     \cellcolor[RGB]{238,238,238}\\
     \cline{1-12}

     
     
     \multirow{6}{*}{10\% (around 8k frames)} & \multirow{4}{*}{PointPillars \cite{lang2019pointpillars}} & - & \textit{From Scratch} & 51.75 & 46.58 & 54.94 & 54.32 & 54.01 & 41.53 & 46.31 &43.88  \\
     
     & & ProCo 
     \cite{yin2022proposalcontrast} & Pre-trained & 54.08 & 49.43 & 57.54 & 56.93 & 56.97 & 45.25 & 47.74 & 46.10 \\ 
     & & \textbf{FAC (Ours)} \cellcolor[RGB]{238,238,238} & Pre-trained \cellcolor[RGB]{238,238,238}  & \textbf{55.16} 
     

     \cellcolor[RGB]{238,238,238} & \textbf{50.51} 
     
     
     \cellcolor[RGB]{238,238,238}  &  \textbf{58.60} 
     
     
     \cellcolor[RGB]{238,238,238}  &  \textbf{57.91}

     
     \cellcolor[RGB]{238,238,238}  &  
     \textbf{57.98} 
     

     \cellcolor[RGB]{238,238,238}  &  \textbf{46.88} 
     
     
     \cellcolor[RGB]{238,238,238}& 
     \textbf{49.19} 
     
     
     \cellcolor[RGB]{238,238,238} & 
     \textbf{47.38} 

     \\

      & & \textbf{FAC++ (Ours)} \cellcolor[RGB]{238,238,238} & Pre-trained \cellcolor[RGB]{238,238,238}
     & \textbf{56.69} 
     \cellcolor[RGB]{238,238,238} 
     & \textbf{51.67} 
     
     \cellcolor[RGB]{238,238,238} 
     & \textbf{59.87}
     \cellcolor[RGB]{238,238,238} 
     & \textbf{59.12} 
     \cellcolor[RGB]{238,238,238} 
     & \textbf{59.07} 
     
     \cellcolor[RGB]{238,238,238}
     & \textbf{47.89}
     
     \cellcolor[RGB]{238,238,238}  
     & \textbf{50.63} 
     
     
     \cellcolor[RGB]{238,238,238}
     & \textbf{48.56} 
     
     \cellcolor[RGB]{238,238,238}\\
      
     \cline{2-12}
     & \multirow{4}{*}{VoxelNet \cite{zhou2018voxelnet}} & - & \textit{From Scratch} & 54.04 & 51.24 & 54.37 & 53.74 & 51.45 & 45.05 & 56.30 & 54.93 \\
      & & ProCo 
     \cite{yin2022proposalcontrast} & Pre-trained & 59.00 & 56.30 & 58.83 & 58.23 & 57.75 & 51.75 & 60.42 & 58.91 \\
     & & \textbf{FAC (Ours)} \cellcolor[RGB]{238,238,238} & Pre-trained \cellcolor[RGB]{238,238,238} &  \textbf{60.16}

     
     \cellcolor[RGB]{238,238,238} &  \textbf{57.23} 
     
     
     \cellcolor[RGB]{238,238,238} &  \textbf{59.90} 
     
     
     \cellcolor[RGB]{238,238,238} &  \textbf{59.71} 
     
     
     \cellcolor[RGB]{238,238,238} &  \textbf{58.85} 
     
     
     \cellcolor[RGB]{238,238,238} & \textbf{52.46} 
     
     
     \cellcolor[RGB]{238,238,238} & \textbf{61.33}  
     

     \cellcolor[RGB]{238,238,238} & 
     \textbf{59.79}

     
     \cellcolor[RGB]{238,238,238} \\

& & \textbf{FAC++ (Ours)} \cellcolor[RGB]{238,238,238} & Pre-trained \cellcolor[RGB]{238,238,238} &  \textbf{61.57}

     \cellcolor[RGB]{238,238,238} &  \textbf{58.16} 
     
     
     \cellcolor[RGB]{238,238,238} &  \textbf{60.99} 
     
     
     \cellcolor[RGB]{238,238,238} &  \textbf{60.89} 
     
     
     \cellcolor[RGB]{238,238,238} &  \textbf{59.98} 
     
     
     \cellcolor[RGB]{238,238,238} & \textbf{53.65} 
     
     
     \cellcolor[RGB]{238,238,238} & \textbf{62.82}  
     

     \cellcolor[RGB]{238,238,238} & 
     \textbf{61.36}

     
     \cellcolor[RGB]{238,238,238} \\

      \bottomrule[0.5mm]
\end{tabular}}
\caption{
Data-efficient \textbf{\textit{3D object detection}} experimental results on Waymo with 1\% and 10\% labeled training data. Similar comparably better experimental results are obtained for KITTI in Table \ref{table_pretrain_kitti} for FAC and FAC++ as compared with the state-of-the-art ProCo~\cite{yin2022proposalcontrast}.   
}
\label{tab_de_waymo}
\vspace{-5.16 mm}
\end{table*}

Following ProCo~\cite{yin2022proposalcontrast} and CSC~\cite{hou2021exploring}, we augment data via random rotation, scaling and flipping, and random point dropout for fair comparisons. We set hyper-parameters $\tau$ in $\mathcal{L}_{Fea}$ and $\mathcal{L}_{Geo}$ at 0.1 following ProCo \cite{yin2022proposalcontrast}, $H$=$f_c$=$m$=20, and the total number of positive/negative pairs as $4096$ in all experiments including detection and segmentation without tuning. In outdoor object detection on Waymo and KITTI \cite{geiger2013vision}, we pre-train the network with Adam~\cite{kingma2014adam} optimizer and follow ProCo \cite{yin2022proposalcontrast} for epoch and batch size setting for fair comparisons with existing works~\cite{liang2021exploring, yin2022proposalcontrast}. In indoor object detection on ScanNet, we follow CSC~\cite{hou2021exploring} to adopt SparseConv~\cite{graham20183d, choy20194d} as the backbone network and VoteNet as the 3D detector, and follow its training settings with the limited number of scene reconstructions \cite{hou2021exploring}. 


\noindent
\textbf{3D Semantic and instance Segmentation.} For 3D segmentation, we strictly follow CSC \cite{hou2021exploring} in the limited reconstruction setting. Specifically, we pre-train on ScanNet and fine-tune pre-trained models on indoor S3DIS, ScanNet and outdoor SemanticKITTI (SK) \cite{behley2019semantickitti}. We use SGD in pre-training with a learning rate of 0.1 and batch size of 32 for 60K steps to ensure fair comparisons with other 3D pre-training methods including CSC~\cite{hou2021exploring} and PCon~\cite{xie2020pointcontrast}. In addition, we test the model pre-trained upon ScanNet for SK to evaluate its learning capacity for outdoor sparse LiDAR point clouds. The only difference is that we fine-tune the model for 320 epochs for SK but 180 epochs for indoor datasets. The longer fine-tuning with SK is because transferring models trained on indoor data to outdoor data takes more time to optimize and converge.


\vspace{-0.02mm}
\vspace{-0.02mm}



\begin{table*}[t]
    \setlength{\abovecaptionskip}{-0.001cm}
    \setlength{\belowcaptionskip}{-0.008cm}
    \centering
    {
    \resizebox{\linewidth}{!}{
        \scalebox{0.96}{\begin{tabular}{l|l||ll|ll|ll|ll|ll}
        \toprule [0.5mm]
           \multirow{2}{*}{Dataset \& Task} & \multirow{2}{*}{Approach} & \multicolumn{2}{c|}{1\%} & \multicolumn{2}{c|}{5\%} & \multicolumn{2}{c|}{10\%} & \multicolumn{2}{c|}{20\%} & \multicolumn{2}{c}{100\%} \\
           &&mAP& AP@50\% & mAP& AP@50\% & mAP& AP@50\% & mAP& AP@50\%  & mAP& AP@50\%\\
        \hline
           \multirow{5}{*}{ScanNet \cite{dai2017scannet} Ins. Seg.}& \textit{From Scratch} & 5.17 & 9.95 & 18.38 & 31.92 & 26.75 &42.76 & 29.39 & 48.12 & 34.53 & 56.97 \\
            & PCon \cite{xie2020pointcontrast}& 7.23 & 12.55 & 19.46 & 35.42 & 27.09 & 43.97 & 30.28 & 49.56 & 37.29 & 58.06 \\
           & CSC \cite{hou2021exploring}& 7.13 & 13.26 & 20.93 & 36.75 & 27.37 &44.93 & 30.62 & 50.61 & 38.69 & 59.43 \\


           & \textbf{FAC (Ours)} \cellcolor[RGB]{238,238,238}& 
          \textbf{13.25 }  \cellcolor[RGB]{238,238,238}         & 
           \textbf{21.95 }            \cellcolor[RGB]{238,238,238}        & 
           \textbf{27.61 }  \cellcolor[RGB]{238,238,238}        & 
           \textbf{44.88 }  \cellcolor[RGB]{238,238,238}        &
           \textbf{30.22 }         \cellcolor[RGB]{238,238,238}        & 
           \textbf{48.23 }    \cellcolor[RGB]{238,238,238}        &
           \textbf{34.57 }   \cellcolor[RGB]{238,238,238}        & 
           \textbf{53.86 }      \cellcolor[RGB]{238,238,238}        & 
           \textbf{40.56 }     \cellcolor[RGB]{238,238,238}       &
           \textbf{60.59 }  \cellcolor[RGB]{238,238,238}\\









   & \textbf{FAC++ (Ours)} \cellcolor[RGB]{238,238,238}& 
          \textbf{14.63}   \cellcolor[RGB]{238,238,238}         & 
           \textbf{22.78}            \cellcolor[RGB]{238,238,238}        & 
           \textbf{28.73} \cellcolor[RGB]{238,238,238}        & 
           \textbf{45.96} \cellcolor[RGB]{238,238,238}        &
           \textbf{36.76}        \cellcolor[RGB]{238,238,238}        & 
           \textbf{52.67}    \cellcolor[RGB]{238,238,238}        &
           \textbf{37.68}  \cellcolor[RGB]{238,238,238}        & 
           \textbf{57.93}     \cellcolor[RGB]{238,238,238}        & 
           \textbf{41.68}     \cellcolor[RGB]{238,238,238}       &
           \textbf{62.67}  \cellcolor[RGB]{238,238,238}\\
           
            \hline

           \multirow{5}{*}{S3DIS \cite{armeni20163d} Ins. Seg.} & \textit{From Scratch} & 9.55 & 13.66 & 20.33 & 30.59 & 25.86 & 36.75 & 28.77 & 40.68 & 40.69 & 59.32 \\
            & PCon \cite{xie2020pointcontrast}& 13.42 & 15.96 & 22.93 & 33.65 & 27.17 & 38.73 & 31.29 & 43.18 & 43.15 & 60.56 \\
           & CSC \cite{hou2021exploring}&  14.66 & 16.70 & 24.91 & 34.23 & 29.77 & 41.08 & 33.59 & 44.77 & 46.25 & 63.48 \\ 
           & \textbf{FAC (Ours)} \cellcolor[RGB]{238,238,238}& 
           \textbf{19.79 } \cellcolor[RGB]{238,238,238} & 
           \textbf{24.62 } \cellcolor[RGB]{238,238,238} & 
           \textbf{28.50 }    \cellcolor[RGB]{238,238,238} & 
           \textbf{39.25 }            \cellcolor[RGB]{238,238,238} & 
           \textbf{35.36 }            \cellcolor[RGB]{238,238,238} & 
           \textbf{45.31}            \cellcolor[RGB]{238,238,238} & 
           \textbf{35.86} 
           
           \cellcolor[RGB]{238,238,238} & 
           \textbf{45.89 
           }            
           
           \cellcolor[RGB]{238,238,238} & 
           \textbf{47.76 }            \cellcolor[RGB]{238,238,238} & 
           \textbf{65.77  }           \cellcolor[RGB]{238,238,238}  \\


 & \textbf{FAC++ (Ours)} \cellcolor[RGB]{238,238,238}& 
           \textbf{21.87 } \cellcolor[RGB]{238,238,238} & 
           \textbf{26.78 } \cellcolor[RGB]{238,238,238} & 
           \textbf{31.74 }  \cellcolor[RGB]{238,238,238} & 
           \textbf{41.56 }            \cellcolor[RGB]{238,238,238} & 
           \textbf{38.89 }            \cellcolor[RGB]{238,238,238} & 
           \textbf{46.57 }            \cellcolor[RGB]{238,238,238} & 
           \textbf{36.92 }            \cellcolor[RGB]{238,238,238} & 
           \textbf{48.25 }            \cellcolor[RGB]{238,238,238} & 
           \textbf{48.98 }            \cellcolor[RGB]{238,238,238} & 
           \textbf{66.87}           \cellcolor[RGB]{238,238,238}  \\ 
        \bottomrule [0.5mm]
        \end{tabular}}}
    \caption{Data efficient \textbf{\textit{3D instance segmentation}} results with the limited number of scene reconstructions on ScanNet~\cite{dai2017scannet} and S3DIS Area-5 validation set with SparseConv~\cite{graham20183d} and PointGroup \cite{jiang2020pointgroup} as the backbone network. FAC clearly outperforms CSC \cite{hou2021exploring} and PCon \cite{xie2020pointcontrast} in all label ratios. The increment is more notable with extremely limited labeled data.}
    \label{tab_ins_seg}
    \vspace{-0.18 mm}
    }
\end{table*}


\begin{table*}[t]
    \setlength{\abovecaptionskip}{-0.00001cm}
    \setlength{\belowcaptionskip}{-0.20007cm}
    \centering
    \resizebox{\linewidth}{!}{
    \begin{tabular}{l||l||l||l||lll||lll||lll}
     \toprule[0.6mm]
     \textbf{Fine-tuning with}  &  & \textbf{Pre-training} & \textbf{mAP}  & \multicolumn{3}{c||}{\textbf{Car}}  & \multicolumn{3}{c||}{\textbf{Pedestrian}} & \multicolumn{3}{c}{\textbf{Cyclist}}\\
     \textbf{Different Label Ratios} & \multirow{-2}{*}{\textbf{3D Detector}} & \textbf{Schedule} & \textbf{(Mod).} & \textbf{Easy} & \textbf{Mod.} & \textbf{Hard} &  \textbf{Easy} & \textbf{Mod.} & \textbf{Hard} & \textbf{Easy} & \textbf{Mod.} & \textbf{Hard}
     \\
     \midrule
     
      
        
     
      
      \multirow{8}{*}{20\% (about 0.74k frames)} & \multirow{4}{*}{PointRCNN~\cite{shi2019pointrcnn}} & \textit{From Scratch} & 63.51 & 88.64 & 75.23 & 72.47 & 55.49 & 48.90 & 42.23 & 85.41 & 66.39 & 61.74 \\
      & & ProCo~\cite{yin2022proposalcontrast}& 66.20 & 88.52 & 77.02 & 72.56 & 58.66 & 51.90 & 44.98 & 90.27 & 69.67 & 65.05 \\
      
      & & \textbf{FAC (Ours)} \cellcolor[RGB]{238,238,238}&\textbf{68.11}
      
      \cellcolor[RGB]{238,238,238} &  \textbf{89.95} 
      
      \cellcolor[RGB]{238,238,238} &  $\textbf{78.75}$ 
      
      
      \cellcolor[RGB]{238,238,238} & \textbf{73.98} 
      
      \cellcolor[RGB]{238,238,238} & \textbf{59.93}
      
      \cellcolor[RGB]{238,238,238} & \textbf{53.98}  \cellcolor[RGB]{238,238,238} & \textbf{46.36}  \cellcolor[RGB]{238,238,238} & \textbf{91.56} \cellcolor[RGB]{238,238,238} & \textbf{72.30} \cellcolor[RGB]{238,238,238} & \textbf{67.88} \cellcolor[RGB]{238,238,238} \\

    & & \textbf{FAC++ (Ours)} \cellcolor[RGB]{238,238,238}&\textbf{69.89} 
      
      \cellcolor[RGB]{238,238,238} &  \textbf{91.26} 
      
      \cellcolor[RGB]{238,238,238} &  $\textbf{80.59}$ 
      
      
      \cellcolor[RGB]{238,238,238} & \textbf{75.36} 
      \cellcolor[RGB]{238,238,238} & \textbf{61.39}
      
      \cellcolor[RGB]{238,238,238} & \textbf{55.57}  \cellcolor[RGB]{238,238,238} & \textbf{47.87}  \cellcolor[RGB]{238,238,238} & \textbf{92.89} \cellcolor[RGB]{238,238,238} & \textbf{73.68} \cellcolor[RGB]{238,238,238} & \textbf{69.96} \cellcolor[RGB]{238,238,238} \\
      
      \cline{2-13}
      
      & \multirow{4}{*}{PV-RCNN~\cite{shi2020pv}} & \textit{From Scratch} & 66.71 & 91.81 & 82.52 & 80.11 & 58.78 & 53.33 & 47.61 & 86.74 & 64.28 & 59.53 \\

      & & ProCo~\cite{yin2022proposalcontrast} & 68.13 & 91.96 & 82.65 & 80.15 & 62.58 & 55.05 & 50.06 & 88.58 & 66.68 & 62.32 \\
       
      & & \textbf{FAC (Ours)} \cellcolor[RGB]{238,238,238}&  \textbf{69.73} 
      
      
      \cellcolor[RGB]{238,238,238} & \textbf{92.87} 
      \cellcolor[RGB]{238,238,238}
      & $\textbf{83.68}$ 
      \cellcolor[RGB]{238,238,238}
      & \textbf{82.32} 
      \cellcolor[RGB]{238,238,238}
      & \textbf{64.15} \cellcolor[RGB]{238,238,238}
      & \textbf{56.78} \cellcolor[RGB]{238,238,238} & \textbf{51.29} \cellcolor[RGB]{238,238,238} & \textbf{89.65} \cellcolor[RGB]{238,238,238}  &\textbf{68.65}
      
      \cellcolor[RGB]{238,238,238} &\textbf{65.63} 
      \\

 & & \textbf{FAC++ (Ours)} \cellcolor[RGB]{238,238,238}&\textbf{71.27} 
      
      \cellcolor[RGB]{238,238,238} &  \textbf{94.79} 
      
      \cellcolor[RGB]{238,238,238} &  $\textbf{85.92}$ 
      
      
      \cellcolor[RGB]{238,238,238} & \textbf{84.99} 
      
      \cellcolor[RGB]{238,238,238} & \textbf{66.87}
      
      \cellcolor[RGB]{238,238,238} & \textbf{57.98}  \cellcolor[RGB]{238,238,238} & \textbf{52.92}  \cellcolor[RGB]{238,238,238} & \textbf{91.39} \cellcolor[RGB]{238,238,238} & \textbf{70.56} \cellcolor[RGB]{238,238,238} & \textbf{67.77} \cellcolor[RGB]{238,238,238} \\
      \midrule

       
     \multirow{10}{*}{100\% (about 3.71k frames)} 
     & \multirow{5}{*}{PointRCNN \cite{shi2019pointrcnn}} & \textit{From Scratch} & 69.45 & 90.02 & 80.56  & 78.02 & 62.59 & 55.66 & 48.69 & 89.87 & 72.12 & 67.52 \\
      
      & & DCon \cite{zhang2021self} & 70.26 & 89.38 & 80.32  & 77.92 & 65.55 & 57.62 & 50.98 & 90.52 & 72.84 & 68.22 \\
      
      & & ProCo \cite{yin2022proposalcontrast} & 70.71 & 89.51 & 80.23 & 77.96 & 66.15 & 58.82 & 52.00 & 91.28 & 73.08 & 68.45 \\
      
      & & \textbf{FAC (Ours)} \cellcolor[RGB]{238,238,238}&  \textbf{71.83} 
      
      \cellcolor[RGB]{238,238,238}&
    
      \textbf{90.53} 

      \cellcolor[RGB]{238,238,238}&
      \textbf{81.29}
      
      \cellcolor[RGB]{238,238,238}&
      \textbf{78.92} 
      
       \cellcolor[RGB]{238,238,238}&
      \textbf{67.23}
      
      \cellcolor[RGB]{238,238,238}&
      \textbf{59.97} 
      
      \cellcolor[RGB]{238,238,238}&
      \textbf{53.10}     
      
      \cellcolor[RGB]{238,238,238} &
      \textbf{92.23} 
      
      \cellcolor[RGB]{238,238,238}&
      \textbf{74.59} 
      
      \cellcolor[RGB]{238,238,238}& 
      \textbf{69.87} \\

       & & \textbf{FAC++ (Ours)} \cellcolor[RGB]{238,238,238}&  \textbf{73.37} 
      
      \cellcolor[RGB]{238,238,238}&
    
      \textbf{92.59} 

      \cellcolor[RGB]{238,238,238}&
      \textbf{82.97}
      
      \cellcolor[RGB]{238,238,238}&
      \textbf{80.59} 
      
       \cellcolor[RGB]{238,238,238}&
      \textbf{69.76}
      
      \cellcolor[RGB]{238,238,238}&
      \textbf{61.99} 
      
      \cellcolor[RGB]{238,238,238}&
      \textbf{55.88}     
      
      \cellcolor[RGB]{238,238,238} &
      \textbf{93.89} 
      
      \cellcolor[RGB]{238,238,238}&
      \textbf{76.38} 
      
      \cellcolor[RGB]{238,238,238}& 
      \textbf{71.96} 
      
      \cellcolor[RGB]{238,238,238} \\  
      \cline{2-13}
      & \multirow{6}{*}{PV-RCNN~\cite{shi2020pv}} & \textit{From Scratch} & 70.57 & - & 84.50 & - & - & 57.06 & - & - & 70.14 & - \\
    
      & & GCC-3D \cite{liang2021exploring} & 71.26 & - &  & - & - & - & - & - & - & - \\      
      & & STRL \cite{huang2021spatio} & 71.46 & - & 84.70 & - & - & 57.80 & - & - & 71.88 & - \\
      & & PCon \cite{xie2020pointcontrast}  & 71.55 & 91.40 & 84.18 & 82.25 & 65.73 & 57.74 & 52.46 & 91.47 & 72.72 & 67.95 \\
      & & ProCo \cite{yin2022proposalcontrast}  & 72.92 & 92.45 & 84.72 & 82.47 & 68.43 & 60.36 & 55.01 & 92.77 & 73.69 & 69.51 \\      
      & & \textbf{FAC (Ours)} \cellcolor[RGB]{238,238,238}& \textbf{73.95} 
      \cellcolor[RGB]{238,238,238} &
      \textbf{92.98} 
      \cellcolor[RGB]{238,238,238} &
      $\textbf{86.33}$ 
      \cellcolor[RGB]{238,238,238}&
      $\textbf{83.82}$ 
      \cellcolor[RGB]{238,238,238}
      &
      $\textbf{69.39}$ 
      \cellcolor[RGB]{238,238,238}
      & 
      $\textbf{61.27}$ 
      \cellcolor[RGB]{238,238,238}
      &
      $\textbf{56.36}$ 
      \cellcolor[RGB]{238,238,238}
      &
      $\textbf{93.75}$ 
      \cellcolor[RGB]{238,238,238}
      &
      $\textbf{74.85}$ 
      \cellcolor[RGB]{238,238,238}
      &
      $\textbf{71.23}$  
      \cellcolor[RGB]{238,238,238}
      \\

      & & \textbf{FAC++ (Ours)} \cellcolor[RGB]{238,238,238}& \textbf{75.77} 
      \cellcolor[RGB]{238,238,238} &
      \textbf{94.49} 
      \cellcolor[RGB]{238,238,238} &
      $\textbf{88.53}$ 
      \cellcolor[RGB]{238,238,238}&
      $\textbf{85.97}$ 
      \cellcolor[RGB]{238,238,238}
      &
      $\textbf{71.92}$ 
      \cellcolor[RGB]{238,238,238}
      & 
      $\textbf{64.21}$ 
      \cellcolor[RGB]{238,238,238}
      &
      $\textbf{59.23}$ 
      \cellcolor[RGB]{238,238,238}
      &
      $\textbf{95.57}$ 
      \cellcolor[RGB]{238,238,238}
      &
      $\textbf{76.39}$ 
      \cellcolor[RGB]{238,238,238}
      &
      $\textbf{73.52}$  
      \cellcolor[RGB]{238,238,238}
      \\
      \bottomrule[0.5mm]
      \end{tabular}}
\caption{
Data-efficient \textbf{\textit{3D object detection}} on KITTI \cite{geiger2013vision}. We pre-train the backbone network of PointRCNN \cite{shi2019pointrcnn} and PV-RCNN~\cite{shi2020pv} on Waymo and transfer to KITTI with 20\% and 100\% annotation ratios in fine-tuning.  FAC as well as FAC++ outperforms the state-of-the-art ProCo \cite{yin2022proposalcontrast} consistently for two settings. \textit{`From Scratch'} denotes the model trained from scratch. All experimental results are averaged over three runs.  }
\label{table_pretrain_kitti}
\end{table*}

\setlength{\tabcolsep}{3.0mm}{
\begin{table}[t]
    \centering
    {
    \resizebox{\linewidth}{!}{
        \scalebox{0.68}{\begin{tabular}{l||l|l|l|l|l}
        \toprule [0.5mm]
           Label Ratio & 10\% & 20\% & 40\% & 80\%  & 100\% \\
        \hline
                \textit{From Scratch} &  0.39 & 4.67 & 22.09 & 33.75 & 35.48  \\
                CSC \cite{hou2021exploring} &8.68  &20.96 & 29.27 & 36.75 & 39.32 \\
                ProCo                   \cite{yin2022proposalcontrast} & 12.64 &  21.87& 31.95 & 37.83  & 40.56 \\
         \rowcolor[RGB]{238,238,238}       \textbf{FAC (Ours)}    
         
        & \textbf{20.96 } 

        
        &
          \textbf{27.35}   

          
        & 
          \textbf{35.93} 

         
        &
          \textbf{39.91} 

          
        &
           \textbf{42.83}     \\

 \rowcolor[RGB]{238,238,238}       \textbf{FAC++ (Ours)}    
         
        & \textbf{22.89} 

        
        &
          \textbf{28.87}   

          
        & 
          \textbf{37.12} 

         
        &
          \textbf{41.23} 

          
        &
           \textbf{44.18}     \\
           
          \bottomrule [0.5mm]
        \end{tabular}}}
    \caption{Data-efficient \textbf{\textit{3D object detection}} average precision (AP\%) with the limited number of scene reconstructions on ScanNet with VoteNet as the backbone network. }
    \label{tab_det}
    }
\vspace{-7.16mm}
\end{table}}

\begin{table}[t] 
    \setlength{\abovecaptionskip}{-0.002cm}
    \setlength{\belowcaptionskip}{-0.002cm}
    \centering
    {
    \resizebox{\linewidth}{!}{
        \begin{tabular}{l|l||l|l|l|l|l}
        \toprule [0.9mm]
           Dataset \& Task & Label Ratio & 1\% & 5\% & 10\% & 20\% & 40\% \\
        \hline
           \multirow{4}{*}{ScanNet Sem. Seg.}& \textit{From Scratch} & 25.65 & 47.06 & 56.72 & 60.93 &  63.72 \\
           & CSC \cite{hou2021exploring} & 29.32 & 49.93 & 59.45 &  64.63 & 68.96  \\
           & \textbf{FAC (Ours)} \cellcolor[RGB]{238,238,238}&\textbf{35.25} \cellcolor[RGB]{238,238,238}  & \textbf{51.95} \cellcolor[RGB]{238,238,238}  & \textbf{61.28} \cellcolor[RGB]{238,238,238} & 
           \textbf{65.84}  \cellcolor[RGB]{238,238,238}& \textbf{69.52}  \cellcolor[RGB]{238,238,238}

           \\
           & \textbf{FAC++ (Ours)} \cellcolor[RGB]{238,238,238} & \textbf{37.71} \cellcolor[RGB]{238,238,238}  & \textbf{53.58} \cellcolor[RGB]{238,238,238}  & \textbf{62.32} \cellcolor[RGB]{238,238,238} & 
           \textbf{66.92}  \cellcolor[RGB]{238,238,238}& \textbf{70.63}  \cellcolor[RGB]{238,238,238}

            \\
           \hline
           \multirow{4}{*}{S3DIS Sem. Seg.}& \textit{From Scratch} &35.75 & 44.38  & 51.86 & 58.72  &  61.83 \\
           & CSC \cite{hou2021exploring} & 36.48 & 45.07  & 52.95 & 59.93  & 62.65 \\
           & \textbf{FAC (Ours)} \cellcolor[RGB]{238,238,238}&\textbf{43.73} \cellcolor[RGB]{238,238,238} & \textbf{49.28} \cellcolor[RGB]{238,238,238}
            & \textbf{54.76} \cellcolor[RGB]{238,238,238} &\textbf{61.05} 
            \cellcolor[RGB]{238,238,238}  &  \textbf{63.22} \cellcolor[RGB]{238,238,238}\\

            & \textbf{FAC++ (Ours)} \cellcolor[RGB]{238,238,238}&\textbf{44.68} \cellcolor[RGB]{238,238,238} & \textbf{50.29} \cellcolor[RGB]{238,238,238}
            & \textbf{55.89} \cellcolor[RGB]{238,238,238} &\textbf{58.92} 
            \cellcolor[RGB]{238,238,238}  &  \textbf{64.36} \cellcolor[RGB]{238,238,238}\\

           \hline
           \multirow{4}{*}{SemanticKITTI \cite{behley2019semantickitti} Sem. Seg.} & \textit{From Scratch} & 28.36 & 33.58 & 46.37 & 50.15 &  54.56\\
           & CSC \cite{hou2021exploring} & 32.78 & 37.55 & 49.62 &  55.67 &  58.89 \\
           & \textbf{FAC (Ours)} \cellcolor[RGB]{238,238,238} & \textbf{39.92} \cellcolor[RGB]{238,238,238}  & \textbf{41.75}  \cellcolor[RGB]{238,238,238}
            & \textbf{52.37} \cellcolor[RGB]{238,238,238} &  \textbf{57.65} \cellcolor[RGB]{238,238,238}&  \textbf{60.17}   \cellcolor[RGB]{238,238,238}\\

          & \textbf{FAC++ (Ours)} \cellcolor[RGB]{238,238,238}&\textbf{41.39} \cellcolor[RGB]{238,238,238} & \textbf{43.13} \cellcolor[RGB]{238,238,238}
            & \textbf{53.26} \cellcolor[RGB]{238,238,238} &\textbf{59.87} 
            \cellcolor[RGB]{238,238,238}  &  \textbf{62.38} \cellcolor[RGB]{238,238,238}\\
            
       \bottomrule [0.9mm]
        \end{tabular}} 
    \caption{Data-efficient \textbf{\textit{3D semantic segmentation}}  (mIoU\%) results with the limited scene reconstructions~\cite{hou2021exploring} on ScanNet, S3DIS, and SemanticKITTI (SK) \cite{behley2019semantickitti} with diverse labelling ratios. 
    }      
    \label{tab_sem_seg}
    }
    \vspace{-8.69998 mm}
\end{table}

\begin{figure*}[t]
    \setlength{\abovecaptionskip}{-0.018cm}
    \setlength{\belowcaptionskip}{-0.118cm}
    \centering
    \includegraphics[scale=0.41808]{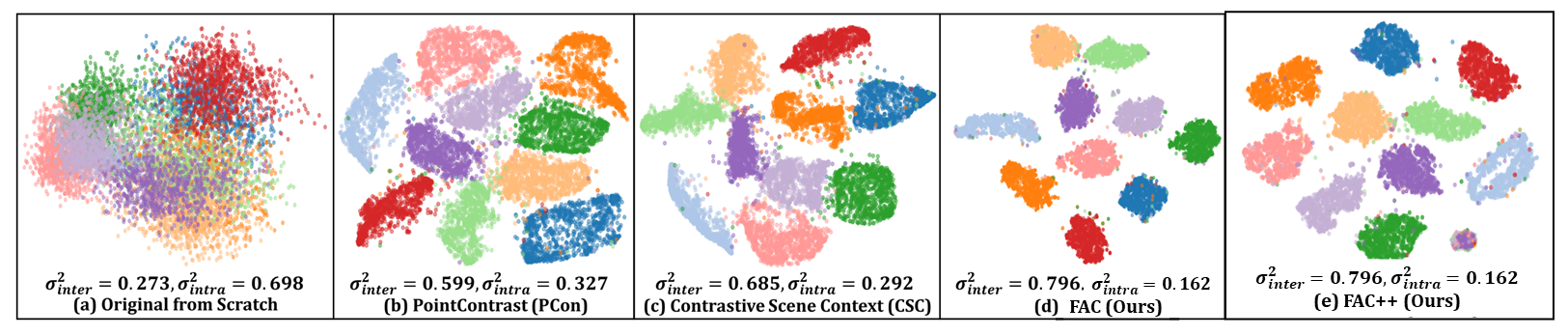}
    \caption{t-SNE~\cite{van2008visualizing} visualization of feature embeddings for SemanticKITTI semantic segmentation fine-tuned with 5\% percent label (ScanNet Pre-trained). Ten classes with the least number of points are shown, where $\sigma^2_{intra}$, $\sigma^2_{inter}$ denote intra- and inter-class variance.  FAC learns a more compact feature space with the smallest intra-class variance and largest inter-class variance as compared with state-of-the-art methods PCon \cite{xie2020pointcontrast}, CSC \cite{hou2021exploring}.
    }
    \label{fig_TSNE}
    \vspace{-0.9388mm}
\end{figure*}


\begin{table*}[t]
    \centering
    \setlength{\abovecaptionskip}{-0.008cm}
    \setlength{\belowcaptionskip}{-0.108cm}
    \resizebox{\linewidth}{!}{
    \scalebox{1.08}{\begin{tabular}{l|lll|llll}
    \toprule [0.5mm]
      Case   & \textit{Sampling} & $\mathcal{L}_{Geo}$ & $\mathcal{L}_{Fea}$ & Sem. mIoU\% (Sc) & Sem. mIoU\% (SK)&  Det. AP@50\% (Sc) & Det. mAP\% (K)\\
    \hline
        \textit{Baseline} &   &  &  &  47.17 & 33.58 & 0.39 & 63.51 \\
        Case 1 & \checkmark& &  & 48.61 & 36.21 & 8.39 &  64.78 \\
        Case 2 & & \checkmark& & 50.62 & 39.87 & 16.46 &  66.36 \\
        Case 3 & & & \checkmark & 50.93 & 40.32 & 17.98 &  66.29 \\
        Case 4 & \checkmark & & \checkmark &  51.34 & 40.56 & 18.57 &  66.72 \\
        Case 5 &  & \checkmark & \checkmark & 51.48 & 40.68 & 18.68 &  67.28 \\
        Case 6 &  \checkmark & \checkmark & & 51.46 & 40.97 & 18.79 &  67.22 \\
    \rowcolor[RGB]{238,238,238}     \textbf{FAC (Full)}    & 
        \checkmark            & 
        \checkmark           & 
        \checkmark         &    
        \textbf{51.95}        &  
          \textbf{41.75}             &    
          \textbf{20.96}       &    
         \textbf{68.11}       \\
        \midrule [0.398mm]

        \textbf{H-FAC} & \checkmark & \checkmark & \checkmark & 51.66 & 41.58  & 20.67 &  67.65 \\
        
    \bottomrule [0.5mm]
    \end{tabular}}}
    \caption{Ablation study of diverse modules of FAC for downstream tasks on ScanNet (Sc) and SemanticKITTI (SK) \cite{behley2019semantickitti}, \& KITTI (K) \cite{geiger2013vision}. 
    \label{tab_ablation}}
    \vspace{-0.2 mm}
\end{table*}

\begin{table*}[t]
    \centering
    \setlength{\abovecaptionskip}{-0.00112cm}
    \setlength{\belowcaptionskip}{-0.00112cm}
    \resizebox{\linewidth}{!}{
    \scalebox{1.08}{\begin{tabular}{l|lll|llll}
    \toprule [0.5mm]
      Case   & \textit{Sampling} & $\mathcal{L}_{Geo}$ & $\mathcal{L}_{Fea}$ & Sem. mIoU\% (Sc) & Sem. mIoU\% (SK)&  Det. AP@50\% (Sc) & Det. mAP\% (K)\\
    \hline
        \textit{Baseline} &   &  &  &  47.87 & 35.08 & 4.78 & 64.68 \\
        Case 1 & \checkmark& &  & 48.93 & 37.47 & 16.21 &  65.52 \\
        Case 2 & & \checkmark& & 51.38 & 39.98 & 17.89 &  66.77 \\
        Case 3 & & & \checkmark & 51.67 & 40.75 & 18.86 &  67.19 \\
        Case 4 & \checkmark & & \checkmark &  52.28 & 41.87 & 19.19 &  66.82 \\
        Case 5 &  & \checkmark & \checkmark & 52.55 & 42.33 & 19.76 &  68.12 \\
        Case 6 &  \checkmark & \checkmark & & 52.89 & 42.56 & 20.87 &  68.61 \\

        \rowcolor[RGB]{238,238,238}     \textbf{FAC++} (Full)   & 
        \checkmark            & 
        \checkmark           & 
        \checkmark         &    
        \textbf{53.58}      &  
          \textbf{43.13}             &    
          \textbf{22.89}       &    
         \textbf{69.96}       \\
        \midrule [0.398mm]
        \textbf{H-FAC++} & \checkmark & \checkmark & \checkmark & 53.42 & 42.97  & 21.76 &  69.22 \\
    \bottomrule [0.5mm]
    \end{tabular}}}
    \caption{Ablation study of diverse modules of FAC++ for downstream scene understanding tasks on ScanNet (Sc) and SemanticKITTI (SK) \cite{behley2019semantickitti}, \& KITTI (K) \cite{geiger2013vision}. 
    \label{tab_ablation_FAC++}}
    \vspace{-1.88 mm}
\end{table*}

\vspace{-6.15mm}

\subsection{Data-efficient Transfer Learning}

\vspace{-3.15mm}

\noindent \textbf{3D Object Detection.} One major target of self-supervised pre-training is more data-efficient transfer learning with less labeled data for fine-tuning. We evaluate data-efficient transfer from Waymo to KITTI \cite{geiger2013vision} as shown in Table \ref{table_pretrain_kitti} and Fig.~\ref{fig_Vis_3D_final}. We can see that FAC outperforms the state-of-the-art consistently. With 20\% labeled data for fine-tuning, FAC achieves comparable performance as training \textit{from scratch} by using 100\% training data, demonstrating its potential in mitigating the dependence on heavy labeling efforts in 3D object detection. Also, as demonstrated in Table~\ref{table_pretrain_kitti}, our proposed FAC++ demonstrates superior performance in the tasks of open-world 3D scene understanding and outperforms previous state-of-the-art by a larger margin. The results reveal that our proposed foreground prompted regional sampling approach can have a significant boost on the final semantic and instance segmentation. These results also to some extent indicate more accurate foreground extraction as well as foreground object awareness can boost the final constrative representations in an effective manner. As Fig.~\ref{fig_activation} shows, FAC has clearly larger activation for inter- and intra-view objects such as vehicles and pedestrians, indicating its learned informative and discriminative representations.


We also study data-efficient learning while performing intra-domain transfer to the Waymo validation set in an extremely label-scarce circumstance with 1\% labels. As Table~\ref{tab_de_waymo} shows, FAC outperforms ProCo \cite{yin2022proposalcontrast} clearly and consistently, demonstrating its potential in reducing data annotations. Moreover, it can also be demonstrated that our proposed FAC++ provides more superior performance gain while trained with less labeled data, demonstrating its label-efficient learning capacity. It can be attributed to that superior foreground-background distinctive representations are learned during the pre-training stage, which boost the final 3D object detection performance. In addition, we conducted experiments for indoor detection on ScanNet. As Table~\ref{tab_det} shows, FAC achieves excellent transfer and improves AP significantly by 20.57\% with 10\% labels compared with \textit{From Scratch}. Also, the improvement is larger when less annotated data is applied. The superior object detection performance is largely attributed to our proposed foreground-grouping aware contrast that leverages informative foreground regions to form the contrast pairs and the adaptive feature contrast that enhances holistic object-level representations.  \\

\begin{table}[t]
    \centering 
    \setlength{\abovecaptionskip}{-0.02cm}
    \setlength{\belowcaptionskip}{-0.1cm}
   \resizebox{\linewidth}{!}{
\begin{tabular}{l|l|l|l|l}
    \toprule
    \multirow{2}{*}{Datasets} & \multirow{2}{*}{Models} & \multicolumn{3}{c}{Few-shot settings} \\
    \cline{3-5}
    && \multicolumn{1}{c|}{B15/N4} &\multicolumn{1}{c|}{B12/N7 } &\multicolumn{1}{c}{B10/N9 }\cr
    \cline{3-5}
    \hline
    \multirow{8}{*}{ScanNet~\cite{dai2017scannet}} & 3DGenZ \cite{cheraghian2020transductive} & 20.6/56.0/12.6& 19.8/35.5/13.3 & 12.0/63.6/6.6\\
     &3DTZSL \cite{michele2021generative} & 10.5/36.7/6.1 & 3.8/36.6/2.0 & 7.8/55.5/4.2 \\
      &LSeg3D \cite{wang20213dioumatch} & 0.0/64.4/0.0 & 0.9/55.7/0.1 & 1.8/68.4/0.9\\
      &PLA without caption \cite{ding2023pla}& 39.7/68.3/28.0 & 24.5/70.0/14.8 & 25.7/75.6/15.5 \\
      &PLA \cite{ding2023pla} & 65.3/68.3/62.4 & 55.3/69.5/45.9 & 
      53.1/76.2/40.8 \\
     & \cellcolor[RGB]{232,232,232} \textbf{FAC \& PLA (Ours)}  & \cellcolor[RGB]{232,232,232} 68.9/70.6/66.8 & \cellcolor[RGB]{232,232,232} 60.3/71.7/58.8 &  \cellcolor[RGB]{232,232,232}
      58.7/77.5/51.6 \\

    & \cellcolor[RGB]{232,232,232} \textbf{FAC \& PLA (Ours)}
   & \cellcolor[RGB]{232,232,232} 70.8/71.9/68.3 & \cellcolor[RGB]{232,232,232} 61.9/72.9/60.5 &  \cellcolor[RGB]{232,232,232}
      59.6/78.8/52.9 \\

    
      & Fully-supervised & 74.5/68.4/79.1 & 73.6/72.0/72.8 &  69.9/75.8/64.9 \\
      \hline  
      \multirow{2}{*}{Datasets} & \multirow{2}{*}{Models} & \multicolumn{3}{c}{Few-shot settings} \\
    \cline{3-5}
    && \multicolumn{1}{c|}{B12/N3} &\multicolumn{1}{c|}{B10/N5 } &\multicolumn{1}{c}{B6/N9}
    \cr
    \cline{3-5}
        \hline
    \multirow{9}{*}{NuScenes~\cite{caesar2020nuscenes}}
     & 3DGenZ \cite{cheraghian2020transductive} & 01.6/53.3/00.8 & 01.9/44.6/01.0 & 01.1/52.6/00.5  \\
     &3DTZSL \cite{michele2021generative} &  01.2/21.0/0.6 & 06.4/17.1/03.9 & 2.61/18.52/03.15\\
     &LSeg-3D~\cite{ding2023pla} & 0.6/74.4/0.3 & 0.0/72.5/0.0 & 2.66/69.72/0.21 \\
      &PLA without caption \cite{ding2023pla}& 25.5/75.8/15.4& 10.7/76.0/05.7 & 8.95/65.83/6.32 \\
      &PLA~\cite{ding2023pla}& 47.7/73.4/35.4 & 24.3/73.1/14.5 & 
      15.63/60.32/12.38 \\
    & \cellcolor[RGB]{232,232,232} \textbf{FAC \& PLA (Ours)} & \cellcolor[RGB]{232,232,232} 52.8/77.3/46.9 & \cellcolor[RGB]{232,232,232} 51.6/79.5/26.8 &\cellcolor[RGB]{232,232,232}  42.5/53.6/58.3\\

     & \cellcolor[RGB]{232,232,232} \textbf{FAC++ \& PLA (Ours)}  & \cellcolor[RGB]{232,232,232} 67.7/79.6/51.8 & \cellcolor[RGB]{232,232,232} 65.9/85.8/47.3 &  \cellcolor[RGB]{232,232,232}
      59.6/77.5/68.9 \\
    
    &  
      Fully-supervised & 73.7/76.6/71.1 & 74.8/76.8/72.8 &  74.6/75.9/72.3 \\
    \bottomrule
    \end{tabular}}
     \caption{Comparisons of the open vocabulary learning performance. It can be demonstrated that our proposed approach provides very superior open-world recognition performance compared with the diverse SOTAs.}  
    \vspace{-2.100mm}
    \label{table_open_world} 
\end{table}

\vspace{-5.8mm}
\noindent \textbf{3D Semantic and Instance Segmentation.}
We first conduct qualitative analysis with point activation maps over the dataset ScanNet. As Fig.~\ref{fig_activation} shows, FAC can find more semantic relationships within and across 3D scenes as compared with the state-of-the-art CSC \cite{hou2021exploring}. This shows that FAC can learn discriminative representations that capture similar features while suppressing distinct ones. As Fig.~\ref{fig_Vis_3D} shows, FAC also produces clearly better instance segmentation as compared with CSC\cite{hou2021exploring}. Specifically, CSC tends to fail to distinguish adjacent instances such as chairs while FAC can handle such challenging cases successfully. 



We also conduct extensive quantitative experiments as shown in the Table \ref{tab_sem_seg}, where we adopt limited labels (\textit{e.g.}, \{1\%, 5\%, 10\%, 20\%\}) in training. We can apparantly see that FAC outperforms the \textit{baseline} \textit{From Scratch} by large margins consistently for both semantic segmentation tasks under different labeling percentages. In addition, FAC outperforms the state-of-the-art CSC~\cite{hou2021exploring} significantly when only 1\% labels are used, demonstrating its capacity in learning informative representations with limited labels. Note FAC achieves more improvements while working with less labeled data. It can also be demonstrated in Table~\ref{tab_sem_seg}, our proposed foreground-prompted regional sampling can also have very beneficial results in the task of semantic segmentation. For example, it improves the performance from 35.25 to 37.71 for the task of semantic segmentation under the 1\% labeled setting. It to some extent validates the better foreground-aware representation can boost the performance of the final data-efficient semantic segmentation. For semantic segmentation over the dataset SK \cite{behley2019semantickitti}, FAC achieves consistent improvement and similar trends with decreasing labeled data.

\vspace{-0.09mm}
\vspace{-0.28mm}

We also study data-efficient learning while performing intra-domain transfer to the Waymo validation set in an extremely label-scarce circumstance with 1\% labels. As Table~\ref{tab_de_waymo} shows, FAC outperforms ProCo \cite{yin2022proposalcontrast} clearly and consistently, demonstrating its potential in reducing data annotations. In addition, we conducted experiments for indoor detection on ScanNet. As Table~\ref{tab_det} shows, FAC achieves excellent transfer and improves AP significantly by 20.57\% with 10\% labels compared with \textit{From Scratch}. Also, the improvement is larger when less annotated data is applied. The superior object detection performance is largely attributed to our foreground-aware contrast that leverages informative foreground regions to form the contrast, and the adaptive feature contrast that enhances the holistic object-level representations. \\

\vspace{-1.688888mm}

The data-efficient open-world recognition results are shown in Figure~\ref{fig_Vis_3D} and Figure~\ref{fig_Vis_3D_final}. It can be demonstrated that better foreground object awareness can be effectively capture by our proposed \textit{FAC++} compared with the state-of-the-art~\cite{ding2023pla}.

 

\vspace{-6.9mm}


\subsection{Open-world 3D scene Understanding}
\vspace{-1.1mm}
We have also extensively evaluated the open-world 3D scene understanding performance of our proposed approaches for the final open-world 3D scene understanding tasks. Specifically, for the open-world 3D scene understanding, we train the model with our proposed FAC for pre-training before we apply the subsequent open-world instance-level 3D scene understanding of PLA~\cite{ding2023pla}. It is demonstrated that our proposed approach has superior performance in terms of open-world 3D scene understanding. As demonstrated in Table~\ref{table_open_world}, our proposed approach achieves superior open-world 3D scene understanding performance. It can be demonstrated that our proposed FAC provides superior performance while combined with the previous state-of-the-art PLA~\cite{ding2023pla}, which demonstrates that the foreground-background distinctive representation is also very fundamental to the final open-world scene understanding performance.  

\vspace{-0.1mm}
In this Subsection, we further evaluate the performance of the open-world recognition capacity of our proposed approach FAC. The results of open-world recognition are demonstrated in Table~\ref{table_open_world}. We have compared our combined FAC \& PLA pre-training with merely adopting the PLA~\cite{ding2023pla} pre-training for establishing the accurate point-language associations. It can be demonstrated that our proposed approach has shown superior performance in terms of open-world recognition. For example, in the setting of B15/N4 our proposed FAC++ has outperformed merely using PLA, which is the previous state-of-the-art by 3.6/2.3/4.4 respectively. The superior performance can be ascribed to that our proposed FAC has demonstrated remarkable performance in establishing foreground-aware feature contrast. We directly use the settings in the PLA~\cite{ding2023pla} and split the categories on ScanNet~\cite{dai2017scannet} and Nuscene~\cite{caesar2020nuscenes} into base and novel categories. It can be demonstrated that our proposed method has superior performance in terms of the open-vocabulary few-shot learning for diverse partitioning of original and novel classes. It can also be demonstrated in Table~\ref{table_open_world} that under diverse spliting of base and novel categories during the data-efficient learning, our porposed FAC++ provides consistent superior performance while conducting 3D scene understanding, demonstrating both its superiority and robustness in terms of open-world 3D recognition while encountered with diverse novel semantic categories and classes.
\vspace{-5.3mm}


\vspace{-1.58mm}
\section{Ablation Study and Analysis}
\vspace{-2mm}
\vspace{-1.9mm}
We perform extensive ablation studies over several key technical designs in FAC. Specifically, we examine the effectiveness of the proposed regional sampling, feature matching network, and the two proposed losses. At the same time, we evaluate the performance of data-efficient learning of our proposed FAC++ as compared with FAC for diverse tasks Lastly, we provide t-SNE visualizations to compare the FAC-learnt feature space with the state-of-the-art. In the ablation studies, we adopt 5\% labels in semantic segmentation experiments, 10\% labels in indoor detection experiments on ScanNet, and 20\% labels in outdoor object detection experiments on KITTI with PointRCNN~\cite{shi2019pointrcnn} as the 3D detector.


\noindent \textbf{Regional Sampling and Feature Matching.}  
Regional sampling samples points in the foreground regions as anchors. The ablation experiment without sampling means that we do not use the foreground sampling and use the random sampled point features to acquire the contrast pairs. Table \ref{tab_ablation} shows related ablation studies as denoted by \textit{Sampling}. We can see that both segmentation and detection deteriorate without \textit{Sampling}, indicating that the foreground regions in over-segmentation may provide important object information while forming contrast. It validates that the proposed regional sampling not only suppresses noises but also mitigates the learning bias towards the background, leading to more informative representations in downstream tasks. In addition, we replace the proposed Siamese correspondence network with Hungarian bipartite matching~\cite{kuhn2005hungarian} (i.e., \textbf{H-FAC}) as shown in Table~\ref{tab_ablation}. We can observe consistent performance drops, indicating that our Siamese correspondence framework can achieve better feature matching and provides well-correlated feature contrast pairs for downstream tasks. More comparisons of matching strategies are reported in the Appendix.

\noindent \textbf{FAC Losses.}  FAC employs a foreground grouping-aware geometric loss $\mathcal{L}_{Geo}$ and a feature loss $\mathcal{L}_{Fea}$ that are critical to its learned representations in various downstream tasks. The geometric loss guides foreground-aware contrast to capture local consistency while the feature loss guides foreground-background distinction. They are complementary and collaborate to learn discriminative representations for downstream tasks. As shown in Table \ref{tab_ablation} cases 4 and 6, including either loss clearly outperforms the \textit{Baseline} as well as the state-of-the-art CSC \cite{hou2021exploring} in segmentation and ProCo \cite{yin2022proposalcontrast} in detection. For example, only including $\mathcal{L}_{Geo}$ (Case 6) achieves 67.22\% and 18.79\% average precision in object detection on KITTI and ScanNet, outperforming ProCo (66.20\% and 12.64\%) by 1.02\% and 6.15\%, respectively as shown in Table \ref{table_pretrain_kitti} and Table \ref{tab_det}. At last, the \textbf{full FAC} in Table \ref{tab_ablation} including both losses learn better representations with the best performance in various downstream tasks.  \\

\vspace{-3.8mm}

\noindent
\textbf{\textbf{FAC++ Ablations:}} The ablation study results of FAC++ are demonstrated in Table \ref{tab_ablation_FAC++}. It is demonstrated that our proposed FAC++ has generally slightly better performance as compared with FAC. The foreground grouping-aware geometric loss $\mathcal{L}_{Geo}$ and the feature loss $\mathcal{L}_{Fea}$ are both very significant for the final scene understanding performance, and dropping either of them will result in significant information loss. As shown in Table \ref{tab_ablation_FAC++} cases 4 and 6, including either loss clearly outperforms the \textit{Baseline} as well as the state-of-the-art CSC~\cite{hou2021exploring} in segmentation and ProCo~\cite{yin2022proposalcontrast} in detection. At the same time, the foreground prompted regional sampling in FAC++ is also very significant for the final scene understanding performance. As validated in Table \ref{tab_ablation_FAC++} case 2, 3, 5, the foreground prompted sampling is of significance to the final downstream performance and removing the foreground sampling results in the performance drop. For example, the performance drops 3.13\% while comparing case 5 with the full \textbf{FAC}. It validates the effectiveness of our proposed foreground prompted sampling in sampling very meaningful and effective foreground-aware feature representations.

\vspace{1mm}
\noindent \textbf{Feature Visualization with t-SNE~\cite{van2008visualizing}}. We employ t-SNE to visualize the feature representations that are learnt for SemanticKITTI \cite{behley2019semantickitti} semantic segmentation task as illustrated in Fig.~\ref{fig_TSNE}. Compared with other contrastive learning approaches such as PCon~\cite{xie2020pointcontrast} and CSC~\cite{hou2021exploring}, FAC learns a more compact and discriminative feature space that can clearly separate features of different semantic classes. As Fig.~\ref{fig_TSNE} shows, the FAC-learnt features have the smallest intra-class variance and largest inter-class variance, demonstrating that the FAC-learnt representations help learn more discriminative features in the downstream task.

\section{Conclusion}

\vspace{-1.6mm}

We propose a \textit{foreground-aware} feature contrast framework (FAC) for 3D unsupervised pre-training. FAC builds better contrastive pairs to produce more geometrically informative and semantically meaningful 3D representations. Specifically, we design a regional sampling technique to promote balanced learning of over-segmented foreground regions and eliminate noisy ones, which facilitates building foreground-aware contrast pairs based on regional correspondence. Moreover, we enhance foreground-background distinction and propose a plug-in-play Siamese correspondence network to find the well-correlated feature contrast pairs within and across views for both the foreground and background segments. Extensive experiments demonstrate the effectiveness as well as the superiority of FAC in terms of both the knowledge transfer and the data efficiency.

\vspace{-3mm}
\bibliographystyle{spmpsci}
{\footnotesize
\bibliography{shortstrings,vggroup,cvww_template,mybib}
}


\clearpage


\begin{figure*}[t]
    \centering
    \includegraphics[scale=0.26998]{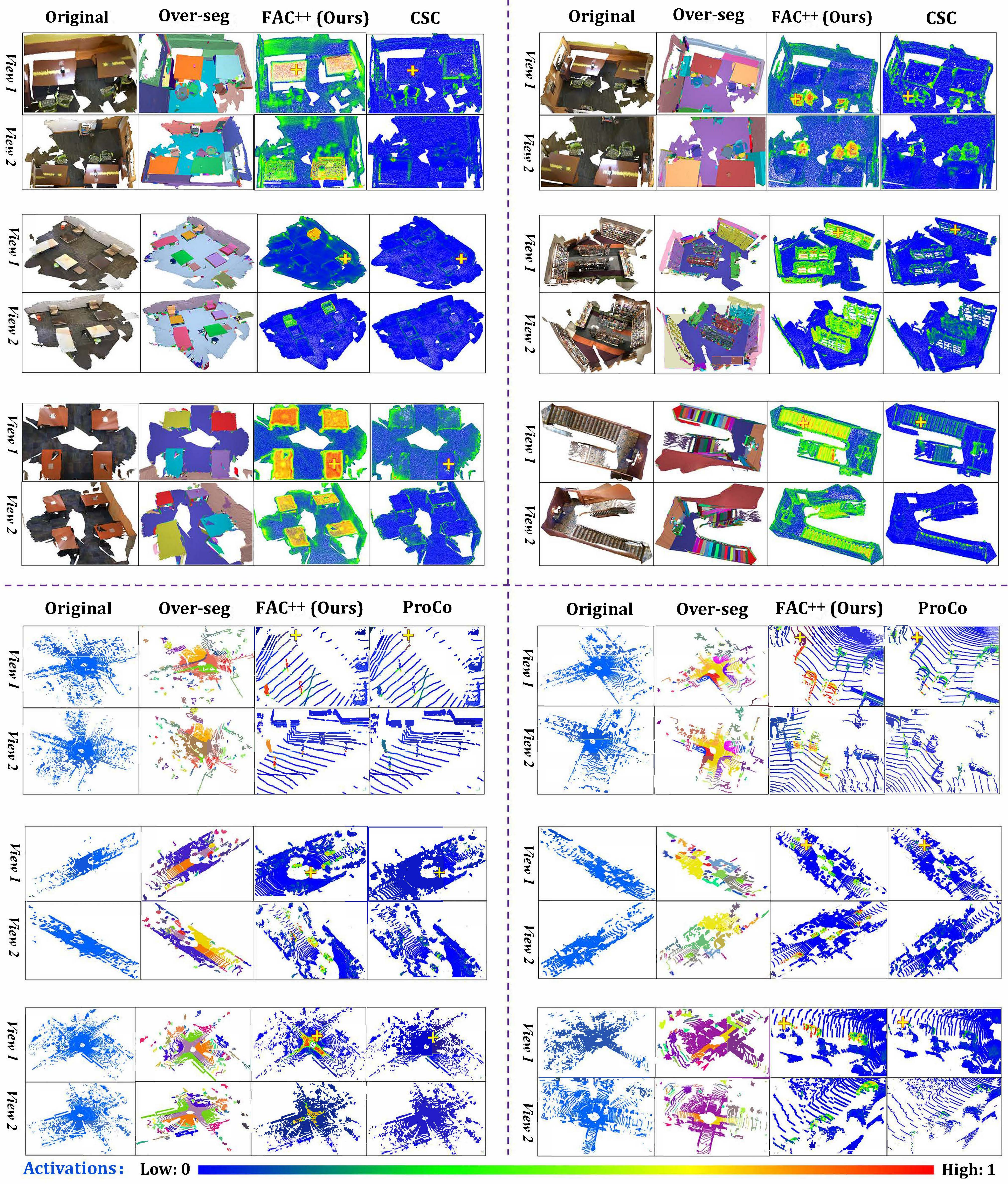}
    \caption{Visualizations of \textit{\textbf{projected point activation maps}} over the indoor ScanNet \cite{dai2017scannet} (Above the purple dash line) and the outdoor KITTI \cite{geiger2013vision} (Below the purple dash line) with respect to the query points highlighted by yellow crosses. The \textit{View 1} and \textit{View 2} in each sample show the intra-view and cross-view correlations, respectively. We compare our proposed FAC++ with the state-of-the-art CSC~\cite{hou2021exploring} on instance segmentation (Above the purple dash line) and ProCo~\cite{yin2022proposalcontrast} on detection (Below the purple dash line). FAC++ clearly captures better feature correlations within and across views as shown in columns 3-4 and columns 7-8 compared with the state-of-the-art approaches CSC~\cite{hou2021exploring} and ProCo~\cite{yin2022proposalcontrast}, respectively.
    }
    \label{fig_activation_supp}
    \vspace{-3mm}
\end{figure*}


\begin{figure*}[t]
    \centering
    \includegraphics[scale=0.36066]{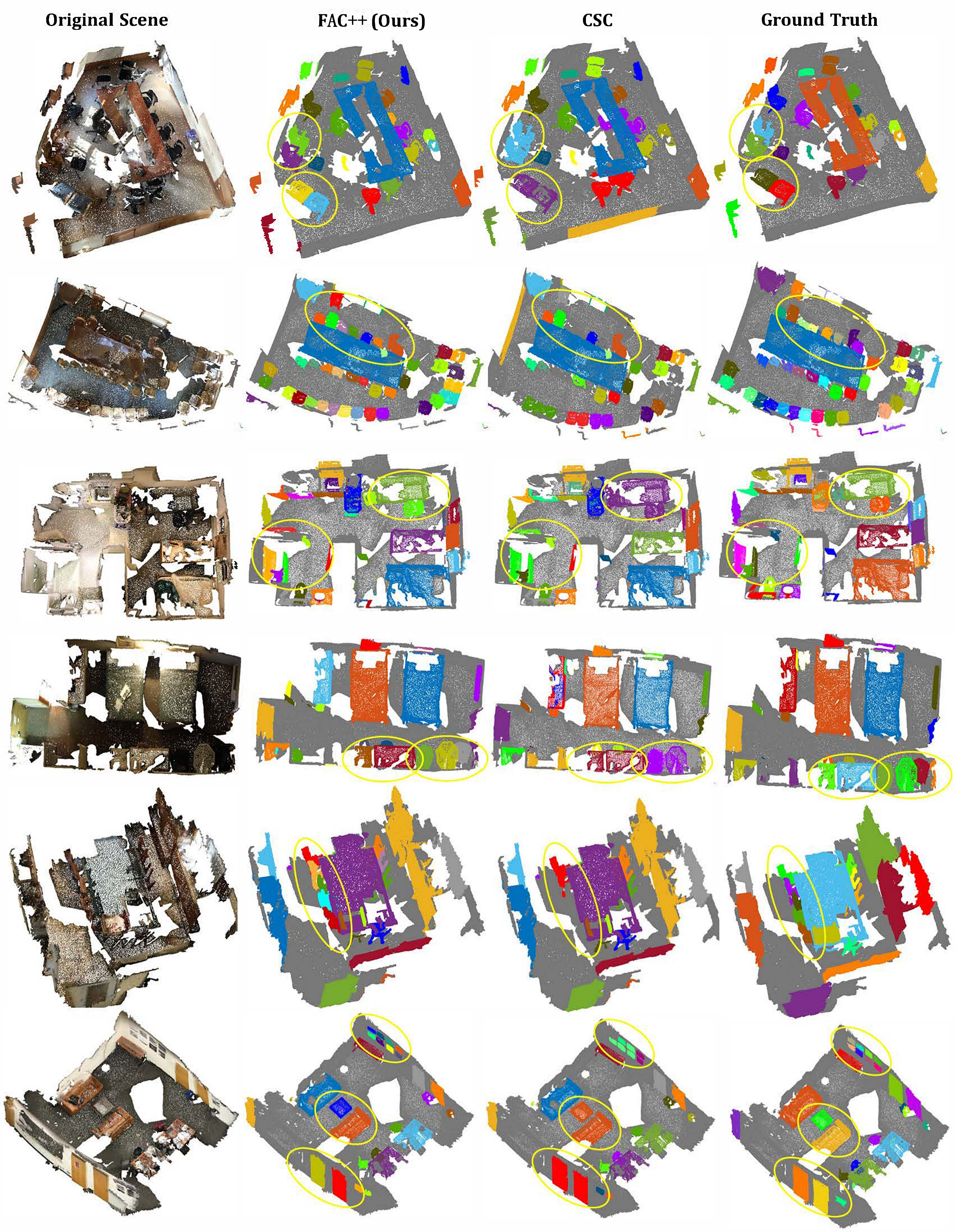}
    \caption{Visualizations of indoor \textit{\textbf{3D instance segmentation}} over ScanNet \cite{dai2017scannet} as fine-tuned with \textit{\textbf{10\% labeled training data}}. Different segmented instances are indicated by different colours. Differences in prediction are highlighted by yellow ellipses.
    }
    \label{fig_3d_results}
\end{figure*}


\begin{figure*}[t]
    \centering
    \includegraphics[scale=0.371]{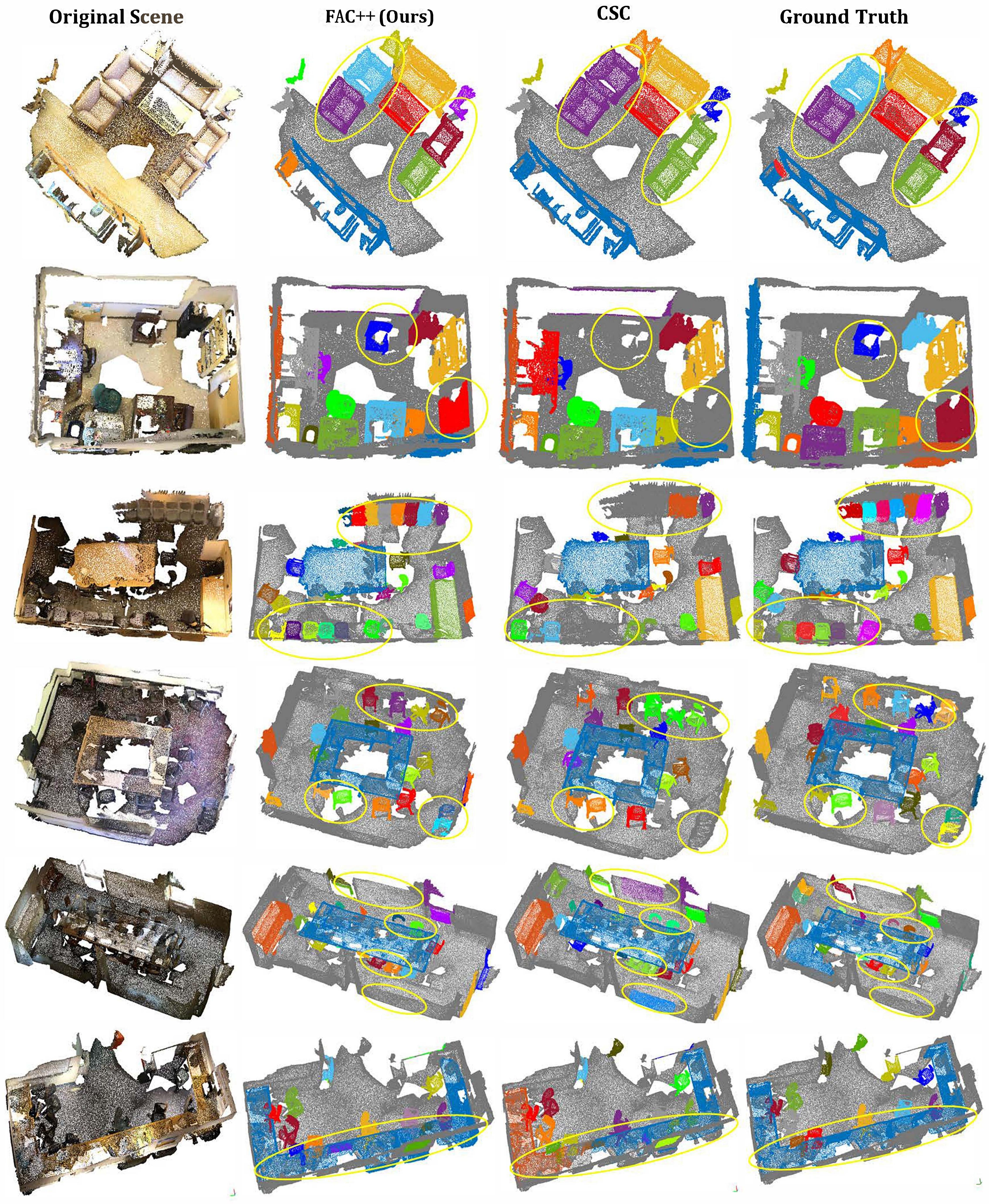}
    \caption{Visualizations of indoor \textit{\textbf{3D instance segmentation}} over ScanNet \cite{dai2017scannet} as fine-tuned with \textit{\textbf{20\% labeled training data}}. Different segmented instances are indicated by different colours. Differences in prediction are highlighted by yellow ellipses.
    }
    \label{fig_3d_results_2}
\end{figure*}

\begin{figure*}[t]
    \centering
    \includegraphics[scale=0.48186]{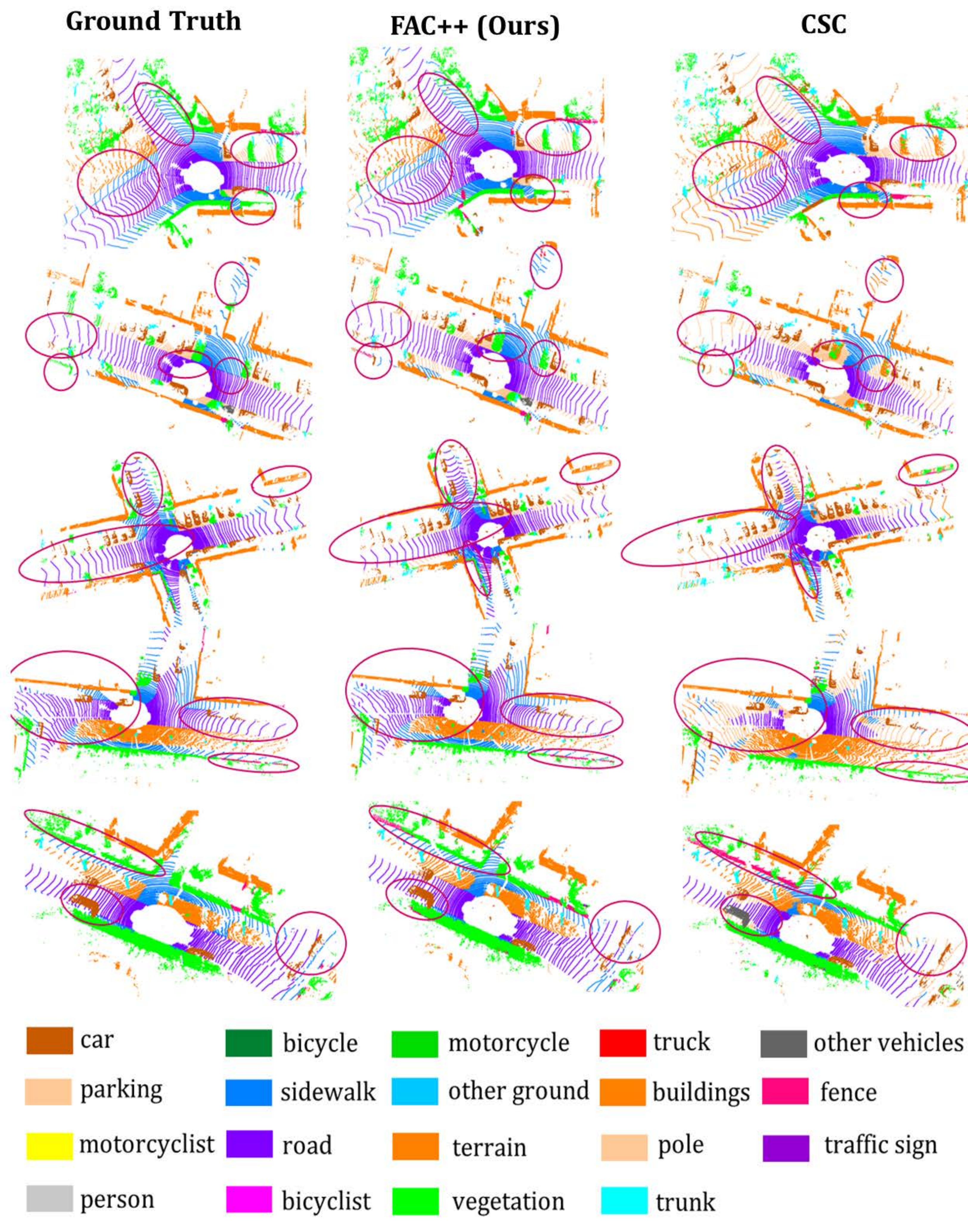}
    \caption{Comparisons of outdoor \textit{\textbf{3D Semantic Segmentation}} Results on SemanticKITTI \cite{behley2019semantickitti} benchmark fine-tuned with \textit{\textbf{10\% labeled training data}} (ScanNet \cite{dai2017scannet} pre-trained). Note that the SemanticKITTI \cite{behley2019semantickitti} has no color channel as input for the task of semantic segmentation. Therefore, we visualize the ground truth without visualizing the original scene (no color channel). Differences in prediction are highlighted by red ellipses.
    }
    \label{fig_skitti_1}
\end{figure*}


\begin{figure*}[t]
    \centering
    \includegraphics[scale=0.4689]{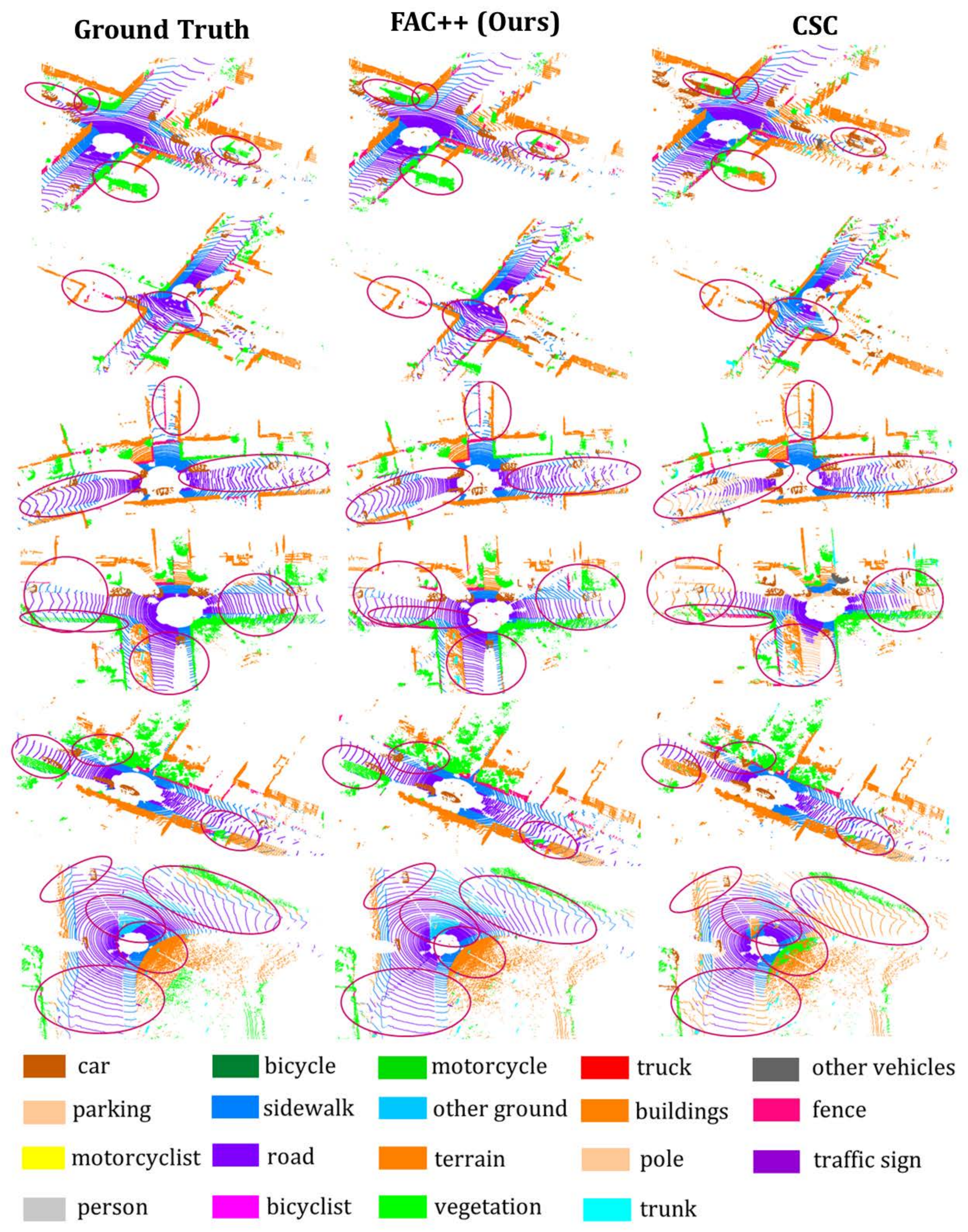}
    \caption{Comparisons of outdoor \textit{\textbf{3D Semantic Segmentation}} Results on SemanticKITTI \cite{behley2019semantickitti} benchmark fine-tuned with \textit{\textbf{20\% labeled training data}} (ScanNet \cite{dai2017scannet} pre-trained). Note that the SemanticKITTI \cite{behley2019semantickitti} has no color channel as input for the task of semantic segmentation. Therefore, we visualize the ground truth without visualizing the original scene (no color channel). Differences in prediction are highlighted by red ellipses.
    }
    \label{fig_skitti_2}
\end{figure*}


\begin{figure*}[htbp!]
    \centering
    \includegraphics[scale=0.6628]{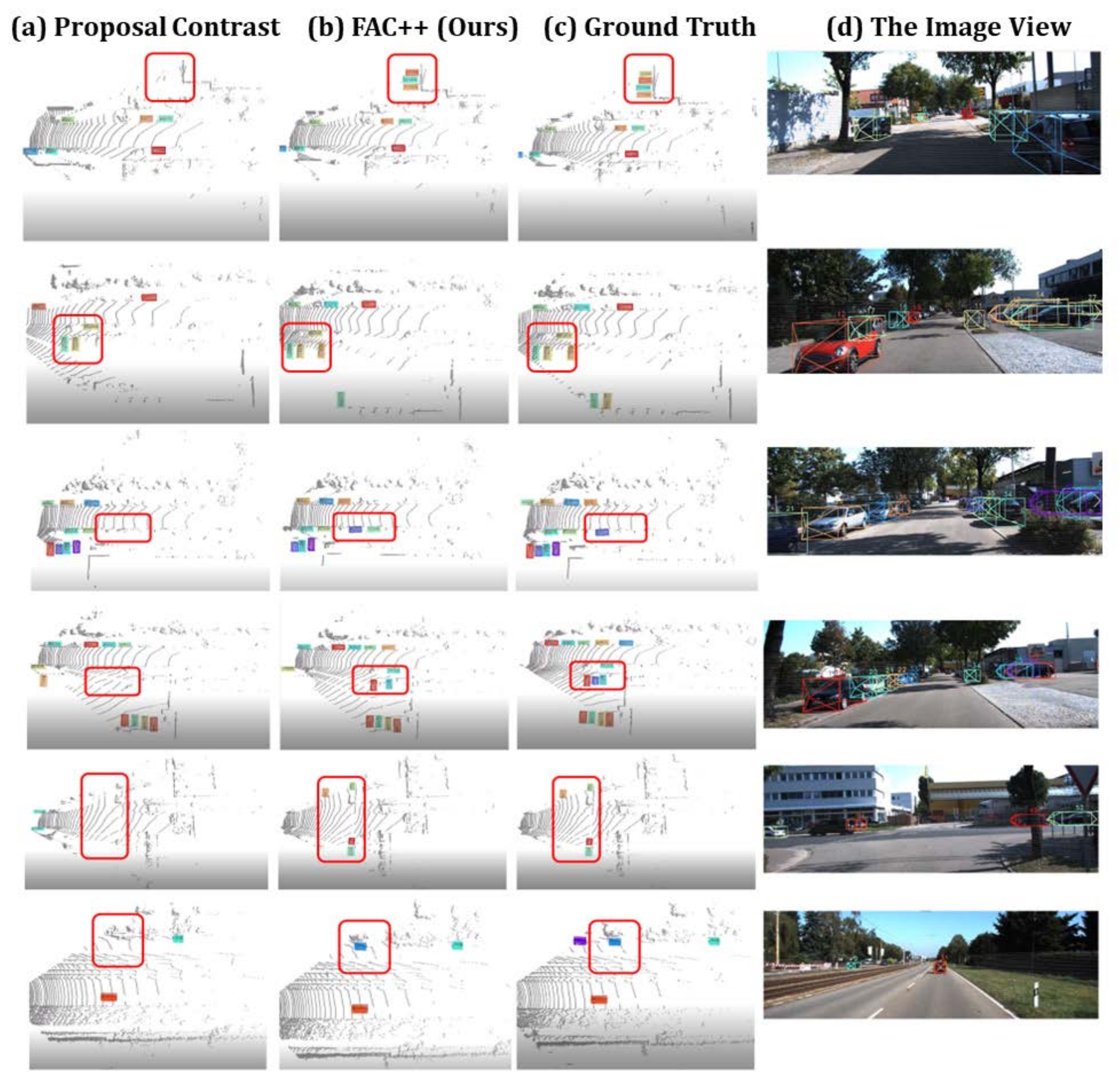}
    \caption{Comparison of \textit{\textbf{3D Object detection}} fine-tuned with \textit{\textbf{20\% labeled training data}} on KITTI \cite{geiger2013vision} benchmark (pre-trained on Waymo \cite{sun2020scalability}) compared with the state-of-the-art approach ProCo \cite{yin2022proposalcontrast}. It can be seen that we can provide more accurate detection results as compared with the state-of-the-art approach ProCo \cite{yin2022proposalcontrast}. Different detected objects are indicated by different colors. Differences in prediction are also highlighted by red rectangles.
    }
    \label{fig_Det}
\end{figure*}

\begin{figure*}[htbp!]
    \centering
    \includegraphics[scale=0.4188]{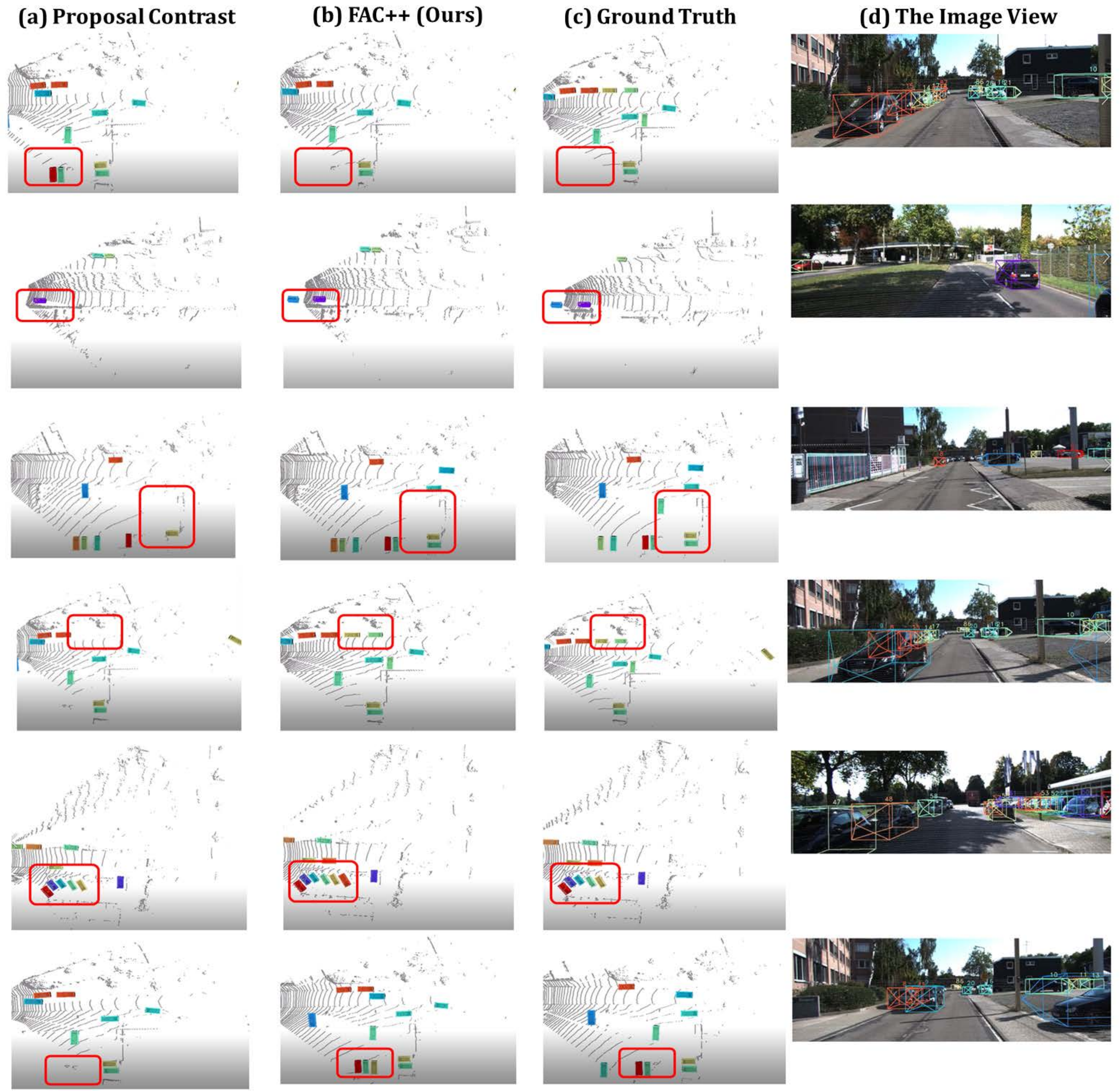}
    \caption{Comparison of \textit{\textbf{3D Object detection}} fine-tuned with \textit{\textbf{50\% labeled training data}} on KITTI \cite{geiger2013vision} benchmark (pre-trained on Waymo \cite{sun2020scalability}) compared with the state-of-the-art approach ProCo \cite{yin2022proposalcontrast}. It can be seen that we can provide more accurate detection results as compared with the state-of-the-art approach ProCo \cite{yin2022proposalcontrast}. Different detected objects are indicated by different colors.  Differences in prediction are also highlighted by red rectangles.
    }
    \label{fig_Det_2}
\end{figure*}

\clearpage

\begin{table*}[t]
\centering
\resizebox{\linewidth}{!}{\scalebox{0.56}{
    \begin{tabular}{l||l|l||l||lll||lll||lll}
     \toprule [0.5mm]
     \rowcolor[RGB]{232,232,232}\textbf{Fine-tuning with}  &  & \textbf{Pre-train.} & \textbf{mAP}  & \multicolumn{3}{c||}{\textbf{Car}}  & \multicolumn{3}{c||}{\textbf{Pedestrian}} & \multicolumn{3}{c}{\textbf{Cyclist}}\\
     \rowcolor[RGB]{232,232,232}\textbf{diverse label ratios} & \multirow{-2}{*}{\textbf{3D Detector}} & \textbf{Schedule} & \textbf{(Mod).} & \textbf{Easy} & \textbf{Mod.} & \textbf{Hard} &  \textbf{Easy} & \textbf{Mod.} & \textbf{Hard} & \textbf{Easy} & \textbf{Mod.} & \textbf{Hard}
     \\
     \hline

      \multirow{8}{*}{20\% (about 0.74k frames)} & \multirow{4}{*}{PointRCNN~\cite{shi2019pointrcnn}} & \textit{From Scratch} & 63.51 & 88.64 & 75.23 & 72.47 & 55.49 & 48.90 & 42.23 & 85.41 & 66.39 & 61.74 \\
      & & ProCo~\cite{yin2022proposalcontrast}& 66.20 & 88.52 & 77.02 & 72.56 & 58.66 & 51.90 & 44.98 & 90.27 & 69.67 & 65.05 \\
      
      & & \textbf{FAC (Ours)} \cellcolor[RGB]{238,238,238}&\textbf{68.11}
      
      \cellcolor[RGB]{238,238,238} &  \textbf{89.95} 
      
      \cellcolor[RGB]{238,238,238} &  $\textbf{78.75}$ 
      
      
      \cellcolor[RGB]{238,238,238} & \textbf{73.98} 
      
      \cellcolor[RGB]{238,238,238} & \textbf{59.93}
      
      \cellcolor[RGB]{238,238,238} & \textbf{53.98}  \cellcolor[RGB]{238,238,238} & \textbf{46.36}  \cellcolor[RGB]{238,238,238} & \textbf{91.56} \cellcolor[RGB]{238,238,238} & \textbf{72.30} \cellcolor[RGB]{238,238,238} & \textbf{67.88} \cellcolor[RGB]{238,238,238} \\

    & & \textbf{FAC++ (Ours)} \cellcolor[RGB]{238,238,238}&\textbf{69.89} 
      
      \cellcolor[RGB]{238,238,238} &  \textbf{91.26} 
      
      \cellcolor[RGB]{238,238,238} &  $\textbf{80.59}$ 
      
      
      \cellcolor[RGB]{238,238,238} & \textbf{75.36} 
      \cellcolor[RGB]{238,238,238} & \textbf{61.39}
      
      \cellcolor[RGB]{238,238,238} & \textbf{55.57}  \cellcolor[RGB]{238,238,238} & \textbf{47.87}  \cellcolor[RGB]{238,238,238} & \textbf{92.89} \cellcolor[RGB]{238,238,238} & \textbf{73.68} \cellcolor[RGB]{238,238,238} & \textbf{69.96} \cellcolor[RGB]{238,238,238} \\
      
      \cline{2-13}
      
         & \multirow{4}{*}{PV-RCNN~\cite{shi2020pv}} & \textit{From Scratch} & 66.71 & 91.81 & 82.52 & 80.11 & 58.78 & 53.33 & 47.61 & 86.74 & 64.28 & 59.53 \\

      & & ProCo~\cite{yin2022proposalcontrast} & 68.13 & 91.96 & 82.65 & 80.15 & 62.58 & 55.05 & 50.06 & 88.58 & 66.68 & 62.32 \\
       
      & & \textbf{FAC (Ours)} \cellcolor[RGB]{238,238,238}&  \textbf{69.73} 
      
      
      \cellcolor[RGB]{238,238,238} & \textbf{92.87} 
      \cellcolor[RGB]{238,238,238}
      & $\textbf{83.68}$ 
      \cellcolor[RGB]{238,238,238}
      & \textbf{82.32} 
      \cellcolor[RGB]{238,238,238}
      & \textbf{64.15} \cellcolor[RGB]{238,238,238}
      & \textbf{56.78} \cellcolor[RGB]{238,238,238} & \textbf{51.29} \cellcolor[RGB]{238,238,238} & \textbf{89.65} \cellcolor[RGB]{238,238,238}  &\textbf{68.65}
      
      \cellcolor[RGB]{238,238,238} &\textbf{65.63} 
      \\

 & & \textbf{FAC++ (Ours)} \cellcolor[RGB]{238,238,238}&\textbf{71.27} 
      
      \cellcolor[RGB]{238,238,238} &  \textbf{94.79} 
      
      \cellcolor[RGB]{238,238,238} &  $\textbf{85.92}$ 
      
      
      \cellcolor[RGB]{238,238,238} & \textbf{84.99} 
      
      \cellcolor[RGB]{238,238,238} & \textbf{66.87}
      
      \cellcolor[RGB]{238,238,238} & \textbf{57.98}  \cellcolor[RGB]{238,238,238} & \textbf{52.92}  \cellcolor[RGB]{238,238,238} & \textbf{91.39} \cellcolor[RGB]{238,238,238} & \textbf{70.56} \cellcolor[RGB]{238,238,238} & \textbf{67.77} \cellcolor[RGB]{238,238,238} \\
      
      \midrule
      \multirow{6}{*}{50\% (about 1.85k frames)} & \multirow{3}{*}{PointRCNN \cite{shi2019pointrcnn}}  & \textit{From Scratch} & 66.73 & 89.12 & 77.85 & 75.36 & 61.82 & 54.58 & 47.90 & 86.30 & 67.76 & 63.26 \\
      
      & & ProCo~\cite{yin2022proposalcontrast} & 69.23 & 89.32 & 79.97 & 77.39 & 62.19 & 54.47 & 46.49 & 91.26 & 73.25 & 68.51 \\
      
      & & \textbf{FAC (Ours)} \cellcolor[RGB]{238,238,238} & 
       \textbf{70.78}  \cellcolor[RGB]{238,238,238}&   \textbf{90.97}  \cellcolor[RGB]{238,238,238} & \cellcolor[RGB]{238,238,238}\textbf{81.52}  & \textbf{78.96}  \cellcolor[RGB]{238,238,238} &   \textbf{63.78} 
       \cellcolor[RGB]{238,238,238} & 
       \textbf{56.01} 
       
       \cellcolor[RGB]{238,238,238} 
       &  \textbf{48.23}
       
       \cellcolor[RGB]{238,238,238} &  $\textbf{92.52}$
      
       \cellcolor[RGB]{238,238,238} &  $\textbf{74.57}$
       
       \cellcolor[RGB]{238,238,238}&  \textbf{69.65}  \cellcolor[RGB]{238,238,238} \\


        & & \textbf{FAC++ (Ours)} \cellcolor[RGB]{238,238,238} & 
       \textbf{71.89}  \cellcolor[RGB]{238,238,238}&   \textbf{91.92}  \cellcolor[RGB]{238,238,238} & \cellcolor[RGB]{238,238,238}\textbf{82.68}  & \textbf{79.99}  \cellcolor[RGB]{238,238,238} &   \textbf{64.87} 
       \cellcolor[RGB]{238,238,238} & 
       \textbf{57.16} 
       \cellcolor[RGB]{238,238,238} 
       &  \textbf{49.38}
       \cellcolor[RGB]{238,238,238} &  $\textbf{93.68}$
       \cellcolor[RGB]{238,238,238} &  $\textbf{75.68}$
       
       \cellcolor[RGB]{238,238,238}&  \textbf{70.69}  \cellcolor[RGB]{238,238,238} \\
      
     \cline{2-13}  
      
      & \multirow{4}{*}{PV-RCNN \cite{shi2020pv}} & \textit{From Scratch} & 69.63 & 91.77 & 82.68 & 81.90 & 63.70 & 57.10 & 52.77 & 89.77 & 69.12 & 64.61 \\
      & & ProCo~\cite{yin2022proposalcontrast} & 71.76 & 92.29 & 82.92 & 82.09 & 65.82 & 59.92 & 55.06 & 91.87 & 72.45 & 67.53 \\ 
      & &  \textbf{FAC (Ours)} \cellcolor[RGB]{238,238,238} & 
       \textbf{73.25}  \cellcolor[RGB]{238,238,238} & 
      \textbf{93.35}  \cellcolor[RGB]{238,238,238} &  
      \textbf{84.39}  \cellcolor[RGB]{238,238,238} &  $\textbf{83.69}$
       \cellcolor[RGB]{238,238,238} &  $\textbf{67.25}$  \cellcolor[RGB]{238,238,238} &  $\textbf{61.45}$  \cellcolor[RGB]{238,238,238} &  $\textbf{56.87}$
      \cellcolor[RGB]{238,238,238} & \textbf{92.17}  \cellcolor[RGB]{238,238,238} &  \textbf{73.96}  \cellcolor[RGB]{238,238,238}&  \textbf{68.98}  \cellcolor[RGB]{238,238,238} \\

 & & \textbf{FAC++ (Ours)} \cellcolor[RGB]{238,238,238} & 
       \textbf{74.53}  \cellcolor[RGB]{238,238,238}&   \textbf{94.78}  \cellcolor[RGB]{238,238,238} & \cellcolor[RGB]{238,238,238}\textbf{85.53}  & \textbf{85.23}  \cellcolor[RGB]{238,238,238} &   \textbf{68.39} 
       \cellcolor[RGB]{238,238,238} & 
       \textbf{62.68} 
       \cellcolor[RGB]{238,238,238} 
       &  \textbf{57.98}
       \cellcolor[RGB]{238,238,238} &  $\textbf{93.53}$
       \cellcolor[RGB]{238,238,238} &  $\textbf{74.97}$
       
       \cellcolor[RGB]{238,238,238}&  \textbf{71.28}  \cellcolor[RGB]{238,238,238} \\
      
      \midrule
       
     \multirow{12}{*}{100\% (about 3.71k frames)} 
     & \multirow{4}{*}{PointRCNN \cite{shi2019pointrcnn}} & \textit{From Scratch} & 69.45 & 90.02 & 80.56  & 78.02 & 62.59 & 55.66 & 48.69 & 89.87 & 72.12 & 67.52 \\
      
      & & DCon \cite{zhang2021self} & 70.26 & 89.38 & 80.32  & 77.92 & 65.55 & 57.62 & 50.98 & 90.52 & 72.84 & 68.22 \\
      
      & & ProCo \cite{yin2022proposalcontrast} & 70.71 & 89.51 & 80.23 & 77.96 & 66.15 & 58.82 & 52.00 & 91.28 & 73.08 & 68.45 \\

 & & \textbf{FAC (Ours)} \cellcolor[RGB]{238,238,238}&  \textbf{71.83} 
      
      \cellcolor[RGB]{238,238,238}&
    
      \textbf{90.53} 

      \cellcolor[RGB]{238,238,238}&
      \textbf{81.29}
      
      \cellcolor[RGB]{238,238,238}&
      \textbf{78.92} 
      
       \cellcolor[RGB]{238,238,238}&
      \textbf{67.23}
      
      \cellcolor[RGB]{238,238,238}&
      \textbf{59.97} 
      
      \cellcolor[RGB]{238,238,238}&
      \textbf{53.10}     
      
      \cellcolor[RGB]{238,238,238} &
      \textbf{92.23} 
      
      \cellcolor[RGB]{238,238,238}&
      \textbf{74.59} 
      
      \cellcolor[RGB]{238,238,238}& 
      \textbf{69.87} \\

       & & \textbf{FAC++ (Ours)} \cellcolor[RGB]{238,238,238}&  \textbf{73.37} 
      
      \cellcolor[RGB]{238,238,238}&
    
      \textbf{92.59} 

      \cellcolor[RGB]{238,238,238}&
      \textbf{82.97}
      
      \cellcolor[RGB]{238,238,238}&
      \textbf{80.59} 
      
       \cellcolor[RGB]{238,238,238}&
      \textbf{69.76}
      
      \cellcolor[RGB]{238,238,238}&
      \textbf{61.99} 
      
      \cellcolor[RGB]{238,238,238}&
      \textbf{55.88}     
      
      \cellcolor[RGB]{238,238,238} &
      \textbf{93.89} 
      
      \cellcolor[RGB]{238,238,238}&
      \textbf{76.38} 
      
      \cellcolor[RGB]{238,238,238}& 
      \textbf{71.96} 
      
      \cellcolor[RGB]{238,238,238} \\  
      
    

      \cline{2-13}
      & \multirow{6}{*}{PV-RCNN \cite{shi2020pv}} & \textit{From Scratch} & 70.57 & - & 84.50 & - & - & 57.06 & - & - & 70.14 & - \\
     
      & & GCC-3D \cite{liang2021exploring} & 71.26 & - &  & - & - & - & - & - & - & - \\
      
      & & STRL \cite{huang2021spatio} & 71.46 & - & 84.70 & - & - & 57.80 & - & - & 71.88 & - \\
      & & PCon \cite{xie2020pointcontrast}  & 71.55 & 91.40 & 84.18 & 82.25 & 65.73 & 57.74 & 52.46 & 91.47 & 72.72 & 67.95 \\
      & & ProCo \cite{yin2022proposalcontrast}  & 72.92 & 92.45 & 84.72 & 82.47 & 68.43 & 60.36 & 55.01 & 92.77 & 73.69 & 69.51 \\

      & & \textbf{FAC (Ours)} \cellcolor[RGB]{238,238,238}& \textbf{73.95} 
      \cellcolor[RGB]{238,238,238} &
      \textbf{92.98} 
      \cellcolor[RGB]{238,238,238} &
      $\textbf{86.33}$ 
      \cellcolor[RGB]{238,238,238}&
      $\textbf{83.82}$ 
      \cellcolor[RGB]{238,238,238}
      &
      $\textbf{69.39}$ 
      \cellcolor[RGB]{238,238,238}
      & 
      $\textbf{61.27}$ 
      \cellcolor[RGB]{238,238,238}
      &
      $\textbf{56.36}$ 
      \cellcolor[RGB]{238,238,238}
      &
      $\textbf{93.75}$ 
      \cellcolor[RGB]{238,238,238}
      &
      $\textbf{74.85}$ 
      \cellcolor[RGB]{238,238,238}
      &
      $\textbf{71.23}$  
      \cellcolor[RGB]{238,238,238}
      \\

      & & \textbf{FAC++ (Ours)} \cellcolor[RGB]{238,238,238}& \textbf{75.77} 
      \cellcolor[RGB]{238,238,238} &
      \textbf{94.49} 
      \cellcolor[RGB]{238,238,238} &
      $\textbf{88.53}$ 
      \cellcolor[RGB]{238,238,238}&
      $\textbf{85.97}$ 
      \cellcolor[RGB]{238,238,238}
      &
      $\textbf{71.92}$ 
      \cellcolor[RGB]{238,238,238}
      & 
      $\textbf{64.21}$ 
      \cellcolor[RGB]{238,238,238}
      &
      $\textbf{59.23}$ 
      \cellcolor[RGB]{238,238,238}
      &
      $\textbf{95.57}$ 
      \cellcolor[RGB]{238,238,238}
      &
      $\textbf{76.39}$ 
      \cellcolor[RGB]{238,238,238}
      &
      $\textbf{73.52}$  
      \cellcolor[RGB]{238,238,238}
      \\
       
     \bottomrule [0.5 mm]
    \end{tabular}}}
\caption{
Data-efficient \textbf{\textit{3D Object Detection}} on KITTI \cite{geiger2013vision}. We pre-train the backbone network of PointRCNN \cite{shi2019pointrcnn} and PV-RCNN \cite{shi2020pv} on Waymo \cite{sun2020scalability} and transfer to KITTI with 20\%, 50\%, and 100\% annotation ratios in fine-tuning.  FAC and FAC++ outperforms the state-of-the-art ProCo \cite{yin2022proposalcontrast} consistently across different settings. \textit{`From Scratch'} denotes the model trained from scratch. }
\label{table_pretrain_kitti_supp}
\vspace{-2mm}
\end{table*}


\begin{table*}[t]
\centering
 \resizebox{\linewidth}{!}{\scalebox{0.838}{
    \begin{tabular}{c|l||c||ll||ll||ll}
        \toprule [0.5mm]
      \rowcolor[RGB]{232,232,232} &  & & \multicolumn{2}{c||}{\textbf{Vehicle}}  & \multicolumn{2}{c||}{\textbf{Pedestrian}} & \multicolumn{2}{c}{\textbf{Cyclist}}\\ 
      
      \rowcolor[RGB]{232,232,232} \multirow{-2}{*}{\textbf{3D Detector}}
       & \multirow{-2}{*}{\textbf{Pre-training Schedule}} & \multirow{-2}{*}{\textbf{Overall AP\%/APH\%}}  & \textbf{AP\%}& \textbf{APH\%} & \textbf{AP\%} & \textbf{APH\%} & \textbf{APH\%} & \textbf{AP\%}
     \\
       \midrule
     
       \multirow{1}{*}{PV-RCNN} & \textit{From Scratch} & 59.84 / 56.23 & 64.99 & 64.38 & 53.80 & 45.14 & 60.61 & 61.35\\
     
       GCC-3D \cite{liang2021exploring} & Pre-trained & 61.30 / 58.18 & 65.65 & 65.10 & 55.54 & 48.02 & 62.72 & 61.43 \\
     
       ProCo \cite{yin2022proposalcontrast} & Pre-trained & 62.62 / 59.28 & 66.04 & 65.47 & 57.58 & 49.51 & 64.23 & 62.86 \\


       \textbf{FAC (Ours)} \cellcolor[RGB]{238,238,238} & Pre-trained \cellcolor[RGB]{238,238,238} &   \textbf{64.57}  / \textbf{61.75}  \cellcolor[RGB]{238,238,238} &   \textbf{68.27}  \cellcolor[RGB]{238,238,238} &   \textbf{67.76}  \cellcolor[RGB]{238,238,238} &    \textbf{59.96}  \cellcolor[RGB]{238,238,238} &  $\textbf{51.27}$   \cellcolor[RGB]{238,238,238} &  \textbf{66.97}  \cellcolor[RGB]{238,238,238} &   \textbf{64.87}  \cellcolor[RGB]{238,238,238} \\

       \textbf{FAC++ (Ours)} \cellcolor[RGB]{238,238,238} & Pre-trained \cellcolor[RGB]{238,238,238} &   \textbf{65.76}  / \textbf{62.89}  \cellcolor[RGB]{238,238,238} &   \textbf{69.53}  \cellcolor[RGB]{238,238,238} &   \textbf{68.95}  \cellcolor[RGB]{238,238,238} &    \textbf{60.97}  \cellcolor[RGB]{238,238,238} &  \textbf{52.38}   \cellcolor[RGB]{238,238,238} &  \textbf{67.88}  \cellcolor[RGB]{238,238,238} &   \textbf{65.92}  \cellcolor[RGB]{238,238,238} \\

        \midrule
       \multirow{1}{*}{CenterPoint \cite{yin2021center}} & \textit{From Scratch} & 63.46 / 60.95 & 61.81 & 61.30 & 63.62 & 57.79 & 64.96 & 63.77 \\
      
        GCC-3D \cite{liang2021exploring} & Pre-trained & 65.29 / 62.79 & 63.97 & 63.47 & 64.23 & 58.47 & 67.68 & 66.44 \\
        ProCo \cite{yin2022proposalcontrast} & Pre-trained & 66.42 / 63.85 & 64.94 & 64.42 & 66.13 & 60.11 & 68.19 & 67.01\\

  \textbf{FAC (Ours)} \cellcolor[RGB]{238,238,238} & Pre-trained \cellcolor[RGB]{238,238,238} & \cellcolor[RGB]{238,238,238} \textbf{68.07}  / \textbf{65.33}  &  \textbf{65.67}  \cellcolor[RGB]{238,238,238}&  \textbf{65.89} \cellcolor[RGB]{238,238,238} &  \textbf{68.90}  \cellcolor[RGB]{238,238,238} &  \textbf{62.70}  \cellcolor[RGB]{238,238,238}&  \textbf{69.06}  \cellcolor[RGB]{238,238,238}&  \textbf{69.27}  \cellcolor[RGB]{238,238,238}\\
        
        \textbf{FAC++ (Ours)} \cellcolor[RGB]{238,238,238} & Pre-trained \cellcolor[RGB]{238,238,238} & \cellcolor[RGB]{238,238,238} \textbf{69.28}  / \textbf{66.86}  &  \textbf{66.95}  \cellcolor[RGB]{238,238,238}&  \textbf{66.92}  \cellcolor[RGB]{238,238,238} &  \textbf{69.97}  \cellcolor[RGB]{238,238,238} &  \textbf{63.89}  \cellcolor[RGB]{238,238,238}&  \textbf{70.69}  \cellcolor[RGB]{238,238,238}&  \textbf{70.53}  \cellcolor[RGB]{238,238,238}\\
       \midrule
       
         CenterPoint-2-Stages (CP2) \cite{yin2021center} & \textit{From Scratch} & 65.29 / 62.47 & 64.70 & 64.11 & 63.26 & 58.46 & 65.93 & 64.85 \\
     
     
        GCC-3D (CP2) \cite{liang2021exploring} & Pre-trained & 67.29 / 64.95 & 66.45 & 65.93 & 66.82 & 61.47 & 68.61 & 67.46 \\
      
        ProCo (CP2) \cite{yin2022proposalcontrast} & Pre-trained & 68.08 / 65.69 &66.98 & 66.48 & 68.15 & 62.61 & 69.04 & 67.97\\
      
        \textbf{FAC (Ours)} \cellcolor[RGB]{238,238,238} & Pre-trained \cellcolor[RGB]{238,238,238} & \cellcolor[RGB]{238,238,238} \textbf{69.95}  / \textbf{67.68}     &  \textbf{68.37} 
      \cellcolor[RGB]{238,238,238}
      &  \textbf{68.56}  \cellcolor[RGB]{238,238,238}   &  \textbf{69.77}  \cellcolor[RGB]{238,238,238} &  \textbf{65.01}  \cellcolor[RGB]{238,238,238} & \textbf{70.55}  \cellcolor[RGB]{238,238,238} &  \textbf{69.38}    \cellcolor[RGB]{238,238,238} \\

        \textbf{FAC++ (Ours)} \cellcolor[RGB]{238,238,238} & Pre-trained \cellcolor[RGB]{238,238,238} & \cellcolor[RGB]{238,238,238} \textbf{71.16}  / \textbf{68.93}  &  \textbf{69.58}  \cellcolor[RGB]{238,238,238}&  \textbf{69.76}  \cellcolor[RGB]{238,238,238} &  \textbf{70.29}  \cellcolor[RGB]{238,238,238} &  \textbf{66.38}  \cellcolor[RGB]{238,238,238}&  \textbf{71.83}  \cellcolor[RGB]{238,238,238}&  \textbf{71.72}  \cellcolor[RGB]{238,238,238}\\
     \bottomrule [0.5mm]
    \end{tabular}}}
\caption{
Data-efficient \textit{\textbf{3D object detection}} on Waymo~\cite{sun2020scalability} with two state-of-the-art 3D object detection backbone networks including PV-RCNN \cite{shi2020pv} and  CenterPoint \cite{yin2021center} fine-tuned with 20\% training labels.  We implement and configure diverse backbone networks with the codebase OpenPCDet \cite{od2020openpcdet}.  Compared with GCC-3D~\cite{liang2021exploring} and ProCo~\cite{yin2022proposalcontrast}, our FAC and FAC++ obtains clear performance gains consistently with different backbone networks. The detectors are trained with 20\% training samples in the training set and evaluated on the validation set.
}
\label{waymo_det_backbone}
\end{table*}
\vspace{-36mm}

\begin{table}[ht]
    \centering
    \scalebox{1.02}{\begin{tabular}{c|l|l}
    \toprule [0.5 mm]
        Pre-training Dataset & Method & Training Time \\
     \midrule [0.3 mm]
     
   \multirow{3}{*}{ ScanNet\cite{dai2017scannet} } 
      &  PCon \cite{xie2020pointcontrast} &  15.25 \\
   &  CSC \cite{hou2021exploring} &  16.13 \\
       & \textbf{FAC (Ours)}  & 16.29 \textbf{\footnotesize{+0.16}}\\
        & \textbf{FAC++ (Ours)}  & 16.28 \textbf{\footnotesize{+0.15}}\\
    
  
    \midrule [0.3 mm]
      \multirow{3}{*}{Waymo \cite{sun2020scalability}}   &  PCon \cite{xie2020pointcontrast} &   20.63 \\
   &  CSC \cite{hou2021exploring} &  21.22 \\
       & \textbf{FAC (Ours)} & 21.37 \textbf{\footnotesize{+0.15}} \\
        & \textbf{FAC++ (Ours)}  & 21.37 \textbf{\footnotesize{+0.15}}\\
    \bottomrule [0.5 mm]
    \end{tabular}}
    \caption{Comparison of the FAC and FAC++ training time with other state-of-the-art 3D pre-training approaches PCon \cite{xie2020pointcontrast} and CSC \cite{hou2021exploring} on different pre-training datasets including ScanNet \cite{dai2017scannet} and Waymo \cite{sun2020scalability}. The unit of the training time is (Miniutes per Epoch).}
    \label{tab:training_time}
\vspace{-3mm}
\end{table}
\clearpage

In this supplementary material, additional experimental results and details that are not included in the main paper due to space limits are provided. We include further parameter analyses and ablation studies testing the robustness of our proposed FAC that are not included due to the space limits. More qualitative and quantitative illustrations are also provided:


\begin{itemize}
     \item Details of our further experimental settings in pre-training including data augmentation and hardware settings (see Section \ref{sec_setting}).

    \item Details of experimental datasets involved in pre-training and testing of our proposed FAC (see Section \ref{subsec:dataset}).
    
    \item Additional quantitative experimental results and analyses are provided (see Section \ref{sec:quan}).
    
    \item Additional qualitative experimental results and analyses are provided (see Section \ref{sec:quali}).

    
    
    \item Details of the further parameter analysis testing the robustness of the proposed FAC are provided. (see Section \ref{sec:effi}).

    \item Future directions of this work (see Section \ref{sec:fu}).
    

\end{itemize}

\vspace{-6mm}

\section{Further Pre-training Experimental Settings}
\label{sec_setting}

\vspace{-1mm}

\subsection{Data Augmentation Details}

\vspace{-1.1mm}
 We utilize four common types of data augmentation to generate augmented two different views in pre-training, including random rotation ([-180\degree, 180\textdegree]) along an arbitrary axis (applied independently for both two views), random scaling ([0.8, 1.2]), random flipping along X-axis or Y-axis, and random point dropout. We follow ProCo \cite{yin2022proposalcontrast} in random point dropout and sample 100k points from the original point cloud for each of the two augmented views. 20k points are chosen from the same indexes to ensure a 20\% overlap for the two augmented views, while the other 80k points are randomly sampled from the remaining point clouds. Our data augmentation strictly follows previous work ProCo \cite{yin2022proposalcontrast} and CSC \cite{hou2021exploring} for fair comparisons with them. Concretely, we follow ProCo \cite{yin2022proposalcontrast} for outdoor 3D object detection on KITTI \cite{geiger2013vision} and Waymo \cite{sun2020scalability} and follow CSC \cite{hou2021exploring} for other experimental cases for data augmentation.

\vspace{-3mm}

 \label{sec:hardware}
\subsection{Hardware Settings }

We next report the hardware used in our experiments. The PCon \cite{xie2020pointcontrast}, ProCo \cite{yin2022proposalcontrast} and CSC \cite{hou2021exploring} use data parallel on eight NVIDIA Tesla V100 GPUs with at least 16 GB GPU memory per card as reported in their papers. Limited by computational resources, we use data parallel on four NVIDIA 2080 Ti GPUs with 11 GB GPU memory per card in all experiments. For experiments in outdoor 3D object detection, we directly report the results of ProCo \cite{yin2022proposalcontrast} in Table 1 and Table 2 of our main paper according to its original paper. It can be seen that FAC still outperforms the state-of-the-art approach ProCo \cite{yin2022proposalcontrast} consistently even if much fewer computational resources are used. For all other experiments, we reimplement the CSC \cite{hou2021exploring}, ProCo 
\cite{yin2022proposalcontrast}, PCon \cite{xie2020pointcontrast} and use the same hardware and experimental settings as our proposed FAC in experiments for a fair comparison in Tables 3, 4, and 5 of our main paper. Specifically, we use data parallel on four NVIDIA 2080 Ti GPUs with 11 GB GPU memory per card.

\vspace{-6mm}

\section{Dataset Details}
\label{subsec:dataset}
\noindent \textbf{S3DIS \cite{armeni20163d}.}  S3DIS is a large indoor point cloud scene understanding dataset across six large-scale indoor areas. The total number of scenes is 271. Area 5 is utilized for testing and other areas are used as the training set. Benefiting from Sparse convolution of Minkowski engine \cite{graham20183d,choy20194d}, we do not partition the 3D scene into small rooms. The S3DIS dataset has more than 215 million points with thirteen semantic classes. It is used to test the effectiveness of the proposed FAC for both indoor semantic segmentation and instance segmentation.

\noindent \textbf{ScanNet-v2 (Sc) \cite{dai2017scannet}.}  ScanNet-v2 is a large-scale and comprehensive 3D indoor scene understanding dataset consisting of 1,513 3D scans. The dataset has been adopted for tasks of semantic segmentation, instance segmentation, and object detection. The dataset is divided into 1,201 scans as the training
set and 312 scans as the validation set. The number of the semantic category is 21 for semantic segmentation. The ScanNet-v2 \cite{dai2017scannet} benchmark is used to test the effectiveness of the proposed FAC for indoor semantic segmentation, instance segmentation as well as indoor object detection. Also, it is used as the pre-training dataset for indoor scene understanding tasks and the outdoor semantic segmentation task on SemanticKITTI \cite{behley2019semantickitti}.

\noindent \textbf{KITTI (K) \cite{geiger2013vision}. } KITTI \cite{geiger2013vision} is a large-scale driving-scene dataset that covers sequential outdoor LiDAR point clouds. The KITTI 3D point cloud object detection dataset consists of 7481 labeled samples. The labeled 3D LiDAR scans are split into the training set with 3,712 scans and the validation set with 3,769 scans. The mean average precision (mAP) with 40 recall positions is typically utilized to evaluate the 3D object detection performance. The 3D IoU (Intersection over Union) thresholds are set as 0.7 for cars and 0.5 for cyclists and pedestrians. The KITTI \cite{geiger2013vision} is used to test the effectiveness of the proposed FAC for outdoor 3D object detection. 

\noindent \textbf{SemanticKITTI (SK) \cite{behley2019semantickitti}. } SemanticKITTI is derived from the above-mentioned KITTI dataset \cite{geiger2013vision} and annotated with point-level semantics. It is made up of more than 43 thousand (43,552) LiDAR scans. It is annotated with nineteen semantic classes. We follow the official split and use sequences 00-10 for training except sequence 08 for validation. The SemanticKITTI \cite{behley2019semantickitti} is used to test the effectiveness of the proposed FAC for outdoor semantic segmentation.

\noindent \textbf{Waymo \cite{sun2020scalability}.} Waymo \cite{sun2020scalability} is a large-scale driving-scene dataset that encompasses 158,361 LiDAR scans from 798 scenes for training and 40,077 LiDAR scans for validation. It is approximately twenty times larger than KITTI \cite{geiger2013vision}. The whole training set (without label) is utilized for pre-training different 3D detection backbone networks. The training set of the Waymo \cite{sun2020scalability} benchmark is used as the pre-training dataset for outdoor 3D object detection. Its validation set is also utilized to test the effectiveness of the proposed FAC for downstream fine-tuning in outdoor 3D object detection.

\vspace{-6mm}

\section{More Quantitative Experiment Results}
\label{sec:quan}
In this Section, we include further quantitative experiments that are not included in the main paper due to space limits for the following three experimental cases:

\subsection{KITTI 3D Object Detection}

       We enrich the experiments of Table 1 in the main paper as shown in Table \ref{table_pretrain_kitti_supp} in this supplementary material. We add the fine-tuning results of data-efficient 3D object detection on KITTI \cite{geiger2013vision} with 50\% labeled training data. From  Table \ref{table_pretrain_kitti_supp}, we can see that although the increments are not as significant as the case when fine-tuned with 20\% labeled training data, FAC can still have a notable boost on data-efficient learning performance when fine-tuned with 50\% labeled training data.  It can also be observed that the improvement is generally more significant as compared with the fine-tuned results with full supervision (100\% labeled training data).
 \subsection{Waymo 3D Object Detection}
    
     We enrich the experiments of Table 2 of the main paper as shown in Table \ref{tab_de_waymo} in this supplementary material. We have added the fine-tuning results of data-efficient 3D object detection on Waymo \cite{sun2020scalability} with more labeled training data including cases with 50\% and 100\% labeled training data (compared with 1\% and 10\% cases). It can be seen that FAC consistently improves the performance with more labeled training data, which further demonstrates the effectiveness of FAC when fine-tuned with the abundant labeled training data.
     
  \subsection{Performance of FAC with Other State-of-the-art 3D Object Detection Backbone Networks}

 We also test the performance of FAC with two state-of-the-art 3D object detection backbone networks including PV-RCNN \cite{shi2020pv} and Centerpoint \cite{yin2021center} as shown in Table \ref{waymo_det_backbone}. We add fine-tuning results of data-efficient 3D object detection on Waymo \cite{sun2020scalability} with 20\% labeled training data. We implement and configure diverse backbone networks with the codebase OpenPCDet \cite{od2020openpcdet}. It can be seen that apart from the 3D backbone network that has been tested in Table \ref{tab_de_waymo}, our proposed FAC also has consistent improvement when pre-trained and fine-tuned with different backbone networks including the state-of-the-art PV-RCNN \cite{shi2020pv} and CenterPoint \cite{yin2021center}, demonstrating the compatibility of FAC while integrated with different 3D object detection backbone networks.

\vspace{-6mm}

\section{More Qualitative Experiment Results}

In this Section, we provide more qualitative experiment results. \textit{First}, we provide more visualizations of the point activation maps to test the learnt representation of our proposed FAC. Concretely, like Fig. 3 in the main paper, visualizations of projected point correlation maps over the indoor ScanNet \cite{dai2017scannet} and the outdoor KITTI \cite{geiger2013vision}  with respect to the query points are provided in Fig.~\ref{fig_activation_supp}.

\textit{Second}, we visualize qualitative data efficient experimental results on various 3D scene understanding tasks with diverse labeling percentages when fine-tuned on downstream tasks including 3D instance segmentation
on ScanNet-v2 \cite{dai2017scannet} as illustrated in Fig. \ref{fig_3d_results} and Fig. \ref{fig_3d_results_2}, 3D semantic segmentation on SemanticKITTI \cite{behley2019semantickitti} as illustrated in Fig. \ref{fig_skitti_1} and Fig. \ref{fig_skitti_2}, and 3D object detection on KITTI \cite{geiger2013vision} as illustrated in Fig. \ref{fig_Det} and Fig. \ref{fig_Det_2}, respectively.

\label{sec:quali}
\subsection{Point Correlation Maps Visualization}

As illustrated in Fig.~\ref{fig_activation}, it is clear that our proposed FAC can effectively find both intra- and inter-view feature correlations of the same semantics compared with the state-of-the-art CSC \cite{hou2021exploring} and ProCo \cite{yin2022proposalcontrast}. For example, as illustrated in Fig.~\ref{fig_activation}, FAC has clearly larger activation for the inter- and intra-view objects of the same semantics as the query point, such as the vehicle, pedestrian, and road. It further demonstrates that FAC learns informative and discriminative representations which capture similar features while suppressing distinct ones.

\subsection{Data-Efficient Instance Segmentation}

The qualitative experimental results of instance segmentation when fine-tuned with 10\% and 20\% labeled training data are shown in Fig. \ref{fig_3d_results} and Fig. \ref{fig_3d_results_2}. It can be seen that in both the above two cases, the state-of-the-art CSC tends to fail to distinguish adjacent instances such as chairs, desks, and sofas, while FAC can handle these challenging cases successfully. 

\vspace{-3.6mm}

    

    
    
    

\subsection{Data-Efficient Semantic Segmentation}
\vspace{-3.6mm}

The qualitative experimental results of semantic segmentation when fine-tuned with 10\% and 20\% labeled training data are shown in Fig. \ref{fig_skitti_1} and Fig. \ref{fig_skitti_2}, respectively. It can be seen that CSC produces many false predictions, while FAC can provide more accurate semantic predictions as compared with the ground truth. It indicates more informative representation is learnt with FAC, which ultimately benefits the downstream semantic segmentation tasks. Also, it demonstrates that the model obtained from indoor pre-training on indoor ScanNet \cite{dai2017scannet} can successfully generalize to outdoor SemanticKITTI \cite{behley2019semantickitti} with data-efficient fine-tuning, which also manifests the generalization capacity of the learnt representation by FAC.

\vspace{-2mm}

\subsection{Data-Efficient Object Detection}
The qualitative experimental results of object detection when fine-tuned with 20\% and 50\% labeled training data are shown in Fig. \ref{fig_Det} and Fig. \ref{fig_Det_2}, respectively. It can be observed that compared with ProCo \cite{yin2022proposalcontrast}, our proposed FAC has clear more accurate predictions in detecting vehicles for outdoor sparse LiDAR point clouds in both 20\% and 50\% labeled training data cases. It further verifies that FAC learns generalized representations that can be applied for both segmentation and detection.

\section{Efficiency Analysis of the Proposed FAC}
\label{sec:effi}

\vspace{-1mm}

To test the efficiency of FAC, we reported the training time of diverse 3D pre-training approaches in Table \ref{tab:training_time}. 
Specifically, we compare with state-of-the-art 3D pre-training approach PCon \cite{xie2020pointcontrast} and CSC \cite{hou2021exploring}. It can be seen that compared with CSC \cite{hou2021exploring}, FAC introduces less than 1\% training overhead on SemanticKITTI \cite{behley2019semantickitti}, and the fine-tuning 
time merely increases by approximately 9 seconds every epoch. The computational overhead mainly comes from the Siamese Correspondence network and the top-$k$ operation. The SCN is light-weighted while the top-$k$ operation has also been implemented with optimal transport in an efficient manner \cite{xie2020differentiable} as illustrated in the main paper. In summary, the efficiency analysis with training time  validates that our proposed FAC merely adds subtle extra computational overhead in pre-training.

\vspace{-1mm}





\section{Future Direction}

\label{sec:fu}

\vspace{-1mm}

In the future, we believe two directions deserve to be further explored to better unleash the potential of 3D unsupervised representation learning. The \textit{first} is constructing large-scale 3D datasets with motion and spatio-temporal statistics for pre-training. The \textit{second} is designing more advanced self-supervised learning techniques leveraging both geometry-aware and semantics-correlated features considering motion and spatio-temporal statistical cues.

\section*{Bibliography}



\noindent
\textbf{Kangcheng Liu} is currently an assistant professor at The Hong Kong University of Science and Technology (Guangzhou), and a joint assistant profesor at The Hong Kong University of Science and Technology. Before that, he was a senior robotics engineer and research fellow at Nanyang Technological University, Singapore, and served as the technical lead of several governmental and industrial projects in Hong Kong and Singapore. He received his B.Eng. degree from the Harbin Institute of Technology and his Ph.D. degree in robotics from the Chinese University of Hong Kong. \\

\noindent
 His interests are robotic systems, vision and graphics, computational design, and manufacturing automation. He has been nominated to serve as the program committee member of international flagship conferences: ICRA, ICPR, ICIP, and CASE. He has published more than 40 international journal and conference papers with wide first-authored publications in TOP-Tier venues including IJCV, IEEE T-MECH, IEEE T-PS, IEEE T-CYB, IEEE T-IE, IEEE T-IP, IEEE T-ITS, CVPR, ECCV, ICCVW, RSS, ICRA, IROS, ACM MM, CBM, and among others as the first author or the corresponding author.
 
 \noindent
 He also served as the reviewer of more than 40 kinds of international journals and conferences such as IJCV, T-PAMI, T-RO, T-CYB, JFR, AIJ, IJRR, CVPR, ICCV, ECCV, TOG, and ACM SIGGRAPH. He is a member of IEEE and ACM. 

 \vspace{12mm}


\noindent
\textbf{
 Xinhu Zheng} received the Ph.D. degree in electrical and computer engineering from the University of Minnesota, Minneapolis, in 2022. He is currently an Assistant Professor with the Hong Kong University of Science and Technology (Guangzhou), and a joint assistant profesor at The Hong Kong University of Science and Technology. His current research interests are data mining in power systems, intelligent transportation system by exploiting different modality of data, leveraging optimization, and machine learning techniques.
 

 \vspace{12mm}

\noindent
\textbf{Kai Tang} is currently a professor at The Hong Kong University of Science and Technology, Guangzhou. He received the B.E. degree in mechanical
engineering from the Nanjing Institute of Technology, China, and the Ph.D. degree in computer engineering from the University of Michigan,
Ann Arbor, USA. \\
From 1990 to 2001, he was at CAD/CAM Software Industry, USA. Since June 2001, he has been
a Faculty Member with The Hong Kong University
of Science and Technology, Hong Kong, where he
is a Professor with the Department of
Mechanical and Aerospace Engineering. His current
research interests include multi-axis machining and additive manufacturing. 

 \vspace{12mm}

\noindent
\textbf{Yong-Jin Liu} received
the B.Eng. degree from Tianjin University, Tianjin,
China, and the M.Phil. and Ph.D. degrees
from The Hong Kong University of Science and
Technology, Hong Kong, China, respectively. He is currently a Professor
with the BNRist, Department of Computer Science and Technology, Tsinghua University, Beijing,
China. His research interests include computational
geometry, computer vision, cognitive computation,
and pattern analysis.  For more information, visit
https://cg.cs.tsinghua.edu.cn/people/~Yongjin/Yongjin.htm

 \vspace{12mm}

\noindent
\textbf{Ming Liu} received the B.A. degree in automation from Tongji University, Shanghai, China, and the Ph.D. degree from the Department of Mechanical Engineering and Process Engineering, ETH Zurich, Zurich, Switzerland. He was a Visiting Scholar with the University of Erlangen-Nuremberg, Erlangen, Germany, and the Fraunhofer Institute of Integrated Systems and Device Technology, Erlangen. He is currently an Associate Professor with the The Hong Kong University of Science and Technology, Guangzhou. His research interests include robotics and autonomous systems, with prominent contributions in autonomous localization and mapping, visual navigation, semantic topological mapping, and environment modeling.

 \vspace{12mm}

\noindent
\textbf{Baoquan Chen} is currently an Endowed Boya professor with the Peking University, where he is the Associate Dean of the School of Artificial Intelligence. He received the MS degree in electronic engineering from Tsinghua University, Beijing, China, and the 2nd MS and PhD degrees in computer science from the State University of New York at Stony Brook, New York, USA.

His research interests generally lie in computer graphics, computer vision, visualization, and human-computer interaction. He has published more than 200 papers in international journals and conferences, including 40+ papers in ACM Transactions on Graphics (TOG)/SIGGRAPH/SIGGRAPH-Asia. Chen serves/served as associate editor of ACM TOG/IEEE Transactions on Visualization and Graphics (TVCG), and has served as conference steering committee member (ACM SIGGRAPH Asia, IEEE VIZ), conference chair (SIGGRAPH Asia 2014, IEEE Visualization 2005), program chair (IEEE Visualization 2004), as well as program committee member of almost all conferences in the visualization and computer graphics fields for numerous times. 
For his contribution to spatial data visualization. He was inducted to IEEE Visualization Academy.

\end{document}